\documentclass[10pt,twocolumn,letterpaper]{article}

\usepackage{wscg}  

\usepackage{graphicx}
\usepackage{amsmath}
\usepackage{amssymb}
\usepackage{booktabs}
\usepackage{comment}
\usepackage{times}
\usepackage{float}
\usepackage{caption}
\usepackage{subcaption}
\usepackage{pgfplots}
\usepackage{tikz-3dplot}
\pgfplotsset{compat=newest}
\usepackage{gensymb}
\usepackage{amssymb}
\usepackage{colortbl}
\usepackage{rotating}
\usepackage[switch]{lineno}

\newcommand\pgfmathsinandcos[3]{%
	\pgfmathsetmacro#1{sin(#3)}%
	\pgfmathsetmacro#2{cos(#3)}%
}
\newcommand\LongitudePlane[3][current plane]{%
	\pgfmathsinandcos\sinEl\cosEl{#2} 
	\pgfmathsinandcos\sint\cost{#3} 
	\tikzset{#1/.style={cm={\cost,\sint*\sinEl,0,\cosEl,(0,0)}}}
}
\newcommand\LatitudePlane[3][current plane]{%
	\pgfmathsinandcos\sinEl\cosEl{#2} 
	\pgfmathsinandcos\sint\cost{#3} 
	\pgfmathsetmacro\yshift{\cosEl*\sint}
	\tikzset{#1/.style={cm={\cost,0,0,\cost*\sinEl,(0,\yshift)}}} %
}

\newcommand\DrawLongitudeCircle[2][1]{
	\LongitudePlane{\angEl}{#2}
	\tikzset{current plane/.prefix style={scale=#1}}
	\pgfmathsetmacro\angVis{atan(sin(#2)*cos(\angEl)/sin(\angEl))} %
	\draw[current plane] (\angVis:1) arc (\angVis:\angVis+180:1);
	\draw[current plane,dashed] (\angVis-180:1) arc (\angVis-180:\angVis:1);
}
\newcommand\DrawLatitudeCircle[2][2]{
	\LatitudePlane{\angEl}{#2}
	\tikzset{current plane/.prefix style={scale=#1}}
	\pgfmathsetmacro\sinVis{sin(#2)/cos(#2)*sin(\angEl)/cos(\angEl)}
	\pgfmathsetmacro\angVis{asin(min(1,max(\sinVis,-1)))}
	\draw[current plane] (\angVis:1) arc (\angVis:-\angVis-180:1);
	\draw[current plane,dashed] (180-\angVis:1) arc (180-\angVis:\angVis:1);
}

\usepackage[pagebackref,breaklinks,colorlinks]{hyperref}

\usepackage[capitalize]{cleveref}
\crefname{section}{Sec.}{Secs.}
\Crefname{section}{Section}{Sections}
\Crefname{table}{Table}{Tables}
\crefname{table}{Tab.}{Tabs.}

\author{
	\parbox{0.25\textwidth}{\centering
		Cl\' ement Hardy\\[1mm]
		Normandie Univ, UNICAEN, GREYC\\
		Caen, France\\[1mm]
		clement.hardy@unicaen.fr
	}
	\hspace{0.05\textwidth}
	\parbox{0.25\textwidth}{\centering
		Yvain Qu\' eau\\[1mm]
		Normandie Univ, CNRS, GREYC\\
		Caen, France\\[1mm]
		yvain.queau@ensicaen.fr
	}
	\hspace{0.05\textwidth}
	\parbox{0.25\textwidth}{\centering
		David Tschumperl\' e\\[1mm]
		Normandie Univ, CNRS, GREYC\\
		Caen, France\\[1mm]
		david.tschumperle@unicaen.fr
	}
}

\begin{document}
\setlength{\parindent}{0pc}

\title{MS-PS: A Multi-Scale Network for Photometric Stereo\\ With a New Comprehensive Training Dataset}

\twocolumn[{\csname @twocolumnfalse\endcsname
	
\maketitle
\begin{abstract}
	
	The photometric stereo (PS) problem consists in reconstructing the 3D-surface of an object,
	thanks to a set of photographs taken under different lighting directions. In this paper, we propose a multi-scale architecture for PS which, combined with a new dataset, yields state-of-the-art results. Our proposed architecture is flexible: it permits to consider a variable number of images as well as variable image size without loss of performance. In addition, we define a set of constraints to allow the generation of a relevant synthetic dataset to train convolutional neural networks for the PS problem. Our proposed dataset is much larger than pre-existing ones, and contains many objects with challenging materials having anisotropic reflectance (e.g.\ metals, glass). We show on publicly available benchmarks that the combination of both these contributions drastically improves the accuracy of the estimated normal field, in comparison with previous state-of-the-art methods.
\end{abstract}

\subsection*{Keywords}
Photometric stereo, 3D-recontruction, normal map estimation, multi-scale achitecture, new dataset
\vspace*{1.0\baselineskip}
}]
\section{Introduction}
\copyrightspace

\noindent Photometric stereo (PS) is a 3D-reconstruction technique that estimates the 3D normal at each point of the surface of an object, using three or more photographs taken from the same viewpoint but with different lighting directions. 
Early works in this field~(e.g.\ \cite{1980}) considered the ideal case of a perfect Lambertian surface. 
However, most images of real world objects exhibit a wide variety of complex lighting effects, which are not well predicted by Lambert's law. Especially, objects' reflectance often includes a specular component, giving a more or less \emph{shiny} appearance to the image surface. Translucent surfaces, such as glass and acrylic, do not respect Lambert's law either. These kind of materials remain in most cases, poorly managed by traditional photometric stereo solutions~\cite{DILI_10}. In order to manage non-Lambertian surfaces, deep learning methods based on convolutional neural networks have recently emerged as the most efficient ones~\cite{DILI_10,dilidataset}. The quality of results obtained by such approaches relies on two main factors: 
\begin{itemize}
	\item[1. ]The architecture of the network, which must ensure a good capacity for generalization on new data, including data with a different size from the training set.\vspace{-0.6em}
	\item[2. ]The quality of the learning dataset, which must be as representative as possible of the diversity of observable light phenomena, for the network to be able to differentiate materials from each other.
\end{itemize}

\vspace*{-1.4em}
\begin{figure}[!ht]
	\centering
	\resizebox{.8\linewidth}{!}{\input{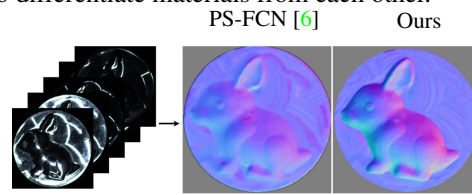}}
	\caption{From a set of images taken under different illumination directions (left), photometric stereo estimates a normal map (right). Our proposed method is particularly efficient when used on challenging anisotropic materials, e.g.\ metal and glass as with this aluminium bunny from~\cite{DILI_10}.}
	\label{fig:diagram_intro}
\end{figure}
\vspace{-1.9em}

\paragraph{Contributions}
\vspace{-0.8em}
Here, we propose a deep learning-based method for the problem of calibrated PS (known lighting direction and intensities), with the following features:\\
\noindent$\bullet$ A multi-scale network architecture for PS, which analyzes the input images simultaneously at different scales;\\
\noindent$\bullet$ A new synthetic training set featuring a wide variety of geometry and non-Lambertian reflectance.\\

\noindent Using these two contributions together, we show that challenging materials with anisotropic reflectance (e.g. metal, glass) can be handled appropriately in the PS problem (Fig.~\ref{fig:diagram_intro}). The underlying core idea is that information over the \textit{whole} image is indeed necessary to infer the 3D normal. Otherwise, complex lighting effects like inter-reflections in metallic objects or sub-surface scattering inside glass cannot be analyzed. On the contrary, our proposed multi-scale architecture takes advantage of all available complex geometric/lighting information and long-distance pixel correlations when inferring the 3D normal map.

\begin{figure*}[!htbp]
	\centering
	\includegraphics[width=0.18\linewidth]{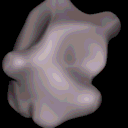}\hspace{0.1cm}
	\includegraphics[width=0.18\linewidth]{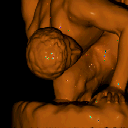}\hspace{0.1cm}
	\includegraphics[width=0.18\linewidth]{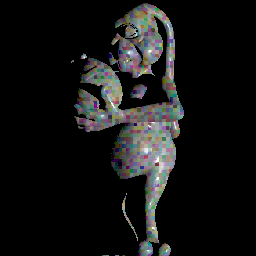}\hspace{0.6cm}
	\includegraphics[width=0.18\linewidth]{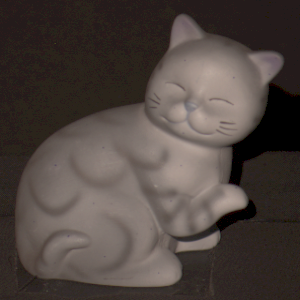}\hspace{0.1cm}
	\includegraphics[width=0.18\linewidth]{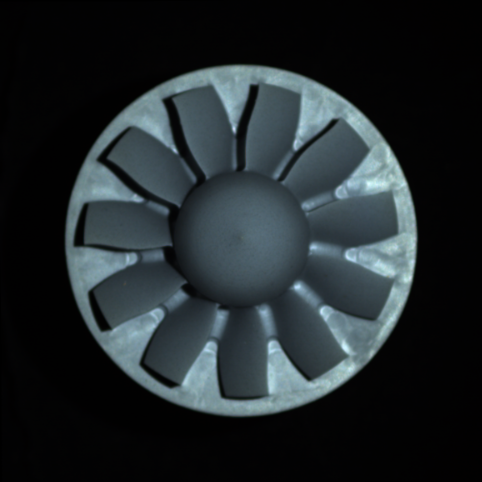}
	\\
	\vspace{0.15cm}
	\subfloat[Blobby~\cite{blobby_dataset}]{\includegraphics[width=0.18\linewidth]{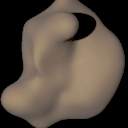}}\hspace{0.1cm}
	\subfloat[Structure~\cite{blobby_dataset}]{\includegraphics[width=0.18\linewidth]{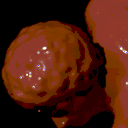}}\hspace{0.1cm}
	\subfloat[CyclePS~\cite{CNN_PS}]{\includegraphics[width=0.18\linewidth]{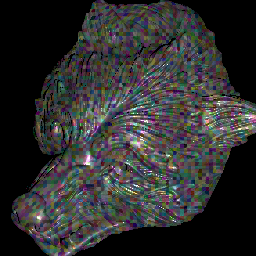}}
	\hspace{0.6cm}
	\subfloat[DiLiGenT~\cite{dilidataset}]{\includegraphics[width=0.18\linewidth]{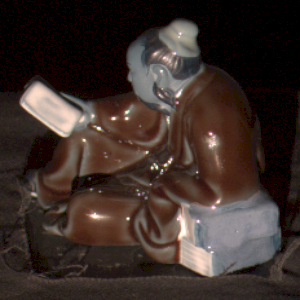}}\hspace{0.1cm}
	\subfloat[DiLiGenT10$^2$~\cite{DILI_10}]{\label{fig:dataset_example_Dili_10}\includegraphics[width=0.18\linewidth]{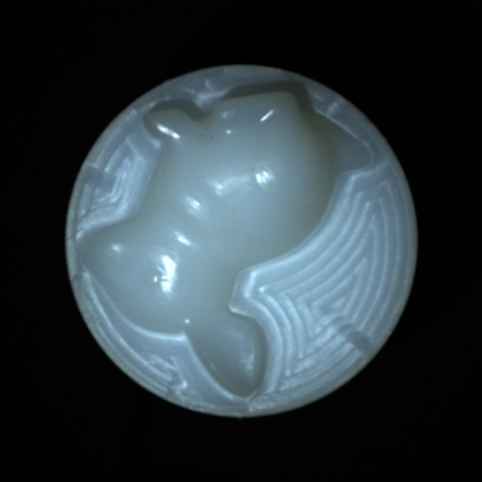}}\\[1em]
	\caption{Samples from existing datasets. The first three \cite{blobby_dataset,CNN_PS} are synthetic ones, used for training the neural networks. Both the last ones \cite{dilidataset,DILI_10} are real-world datasets used for benchmarking. Our proposed multi-scale architecture is evaluated on both benchmarking datasets, and trained on a new synthetic training set, which contains much more objects with \emph{non-Lambertian} reflectance.}
	\label{fig:datasets}
\end{figure*}

\vspace*{-0.3em}
\section{Related Work}
\label{sec:relatedworks}

Deep learning techniques for photometric stereo are all based on the use of Convolutional Neural Networks (CNN). Typically, a fully CNN architecture requires a fixed number of input images. However, in photometric stereo, the number of images depends on the acquisition procedure. To avoid having to train a different network model for each possible number of input images, two alternatives have been considered in the literature.

\vspace*{-1em}
\paragraph{Observation map VS pooling}
\vspace*{-0.5em}
The first alternative consists in using an observation map~\cite{CNN_PS,ICIP_2022,minify,PX_NET,spline_net}, which projects all observations of each pixel under different illuminations into a fixed-size space - typically a sampled hemisphere. Therefore, an observation map makes a fixed-size summary of the information contained in a variable-size set of images. However, the spatial information (intra image) is lost, and the performance drops when the number of input images is small (typically,  $<$10)~\cite{PS_transformer}.\\
The second alternative rather resorts to specific pooling modules~\cite{SDPS_net,PS_FCN,9376632,pay_attention,9069410}, which aggregate the different features of each image extracted by previous convolution layers. This allows to obtain fixed-size image features from a variable number of input images. Different pooling methods can be considered. It is shown in~\cite{PS_FCN} that max pooling performs better than average pooling as soon as the number of images exceeds~$16$. The latter tends to over-smooth the salient features and to be too sensitive to the regions of images with little interest, although a max pooling can also sometimes ignore a large proportion of the features extracted~\cite{9301860}. Still, in contrast to the observation map approach, pooling methods pay attention to intra image information, despite using less the variations of pixel values across the images.

\vspace*{-1em}
\paragraph{Architectural variants}
\vspace*{-0.5em}
To overcome the drawbacks of both these approaches, Yao et al.~\cite{GPS_net} introduced a graph method called GPS-NET. It first aggregates the inter-image information by using a graph structure, and then uses a CNN to predict a 3D normal map. This graph structure therefore allows to preserve the spatial information. More recently, Ikehata~\cite{PS_transformer} proposed a dual-branch transformer (PS-transformer). One branch takes as input the pixels under different illuminations to get the inter-information, the other branch processes the images to get the spatial one. The features extracted are then aggregated, and a CNN finally gives the 3D normal map. However, as mentioned in~\cite{PS_transformer} transformers are not particularly suitable for dense problems (in our case, a large number of input images).

\noindent In this paper, we rather consider the pooling-based scheme from~\cite{PS_FCN} as a baseline model, and broaden it to a multi-scale architecture. Multi-scale architecture for photometric stereo has already been used, e.g., by  Lichy et al. in the context of directional lighting with few images (no more than 6 images in inputs)~\cite{shape_and_material}, or for near (non-parallel) lighting~\cite{fast_near}. On the contrary, we design our method to handle the directional lighting case with a large number of input images (e.g. 96 images). 


\vspace*{-1em}
\paragraph{Existing training datasets}
\vspace*{-0.5em}
Regardless of its architecture, a neural network needs to be trained on a proper dataset to perform well. In practice though, it is very difficult to acquire a large dataset of real images with 3D ground truths of photographed objects. For this reason, deep photometric stereo networks proposed in the literature often rely on training datasets of synthetic 3D objects, notably the \emph{Blobby} and \emph{Structure} datasets introduced in~\cite{PS_FCN}, and \emph{CyclePS} in~\cite{CNN_PS}. 

\begin{figure}[!htpb]
	\captionsetup[subfigure]{justification=centering}
	\centering
	\begin{subfigure}{0.33\linewidth}
		\resizebox{1.1\linewidth}{!}{
			\tdplotsetmaincoords{180}{180} 
\begin{tikzpicture}[tdplot_main_coords]
\def\R{1.3} 
\def\Raxes{\R*1.3}
\def\angEl{10} 
\foreach \t in {-80,-70,...,80} { \DrawLatitudeCircle[\R]{\t} }
\foreach \t in {5,15,...,175} { \DrawLongitudeCircle[\R]{\t} }

\draw[thick,->] (0,0,0) -- (\Raxes,0,0) node[anchor=north east]{x};
\draw[-stealth] (0,0,0) -- (0,\Raxes,0) node[anchor=north east]{y};
\draw[-stealth] (0,0,0) -- (0,0,\Raxes*1) node[anchor=north east]{z};
\draw[dashed, gray] (0,0,0) -- (-\Raxes,0,0);
\draw[dashed, gray] (0,0,0) -- (0,-\Raxes,0);

\foreach \i/\x/\y/\z/\u in {'1/-0.03280081/0.41611025/0.90872238/1',
 '1/-0.03839991/0.31049925/0.9497977/1',
 '1/-0.04399823/0.19019233/0.98076044/1',
 '1/-0.04919816/0.05869781/0.99706274/1',
 '1/-0.05360061/-0.07800089/0.99551134/1',
 '1/-0.05710194/-0.21230722/0.97553319/1',
 '1/-0.05939919/-0.33739538/0.93948714/1',
 '1/-0.06059872/-0.44839055/0.8917812/1',
 '1/-0.15270716/0.4160195/0.89644202/1',
 '1/-0.16449474/0.31129005/0.93597008/1',
 '1/-0.17500445/0.19220488/0.96562453/1',
 '1/-0.18329657/0.06229884/0.98108166/1',
 '1/-0.18449839/-0.43989616/0.87889232/1',
 '1/-0.18849605/-0.07269848/0.97937948/1',
 '1/-0.18870114/-0.32950199/0.92510557/1',
 '1/-0.1901994/-0.20549935/0.95999698/1',
 '1/-0.26670649/0.41020998/0.87212122/1',
 '1/-0.28359251/0.30729189/0.90837601/1',
 '1/-0.2981084/0.19100538/0.93522635/1',
 '1/-0.30139786/-0.42509698/0.85349393/1',
 '1/-0.30880699/0.06470146/0.94892147/1',
 '1/-0.31020115/-0.31660118/0.89640333/1',
 '1/-0.31450962/-0.06640203/0.94692897/1',
 '1/-0.31489186/-0.19549495/0.92877599/1',
 '1/-0.3707838/0.39958255/0.83836338/1',
 '1/-0.39141794/0.29941372/0.87013988/1',
 '1/-0.40709586/-0.40559588/0.81839168/1',
 '1/-0.40860354/0.18720162/0.89330773/1',
 '1/-0.41919869/-0.29999906/0.85689732/1',
 '1/-0.42090385/0.0659006/0.90470827/1',
 '1/-0.42618633/-0.18329412/0.88587159/1',
 '1/-0.42691512/-0.05950211/0.90233197/1',
 '1/-0.46289771/0.38549809/0.79819605/1',
 '1/-0.48561334/0.28870793/0.82512267/1',
 '1/-0.499596/-0.38319693/0.77689378/1',
 '1/-0.50428312/0.18139393/0.84427174/1',
 '1/-0.51360334/-0.28130183/0.81060526/1',
 '1/-0.51720021/0.06620003/0.85330035/1',
 '1/-0.52190667/-0.16990217/0.83591069/1',
 '1/-0.52326874/-0.05249686/0.85054918/1',
 '1/-0.54241965/0.36921337/0.75462733/1',
 '1/-0.56588359/0.27639198/0.77677747/1',
 '1/-0.57853599/-0.35932235/0.73224555/1',
 '1/-0.58486953/0.17439091/0.79215873/1',
 '1/-0.59329349/-0.26189713/0.76119165/1',
 '1/-0.59781529/0.06590169/0.79892043/1',
 '1/-0.60202327/-0.15630604/0.78303027/1',
 '1/-0.6037003/-0.04570002/0.7959004/1',
 '1/0.03440055/0.41730662/0.90811441/1',
 '1/0.03870043/0.30950341/0.95011048/1',
 '1/0.04279892/0.18779527/0.98127529/1',
 '1/0.04640163/0.05630197/0.99733494/1',
 '1/0.04919919/-0.07869871/0.99568366/1',
 '1/0.05109826/-0.20999284/0.97636672/1',
 '1/0.05199786/-0.33138635/0.9420612/1',
 '1/0.05220076/-0.43870637/0.89711304/1',
 '1/0.1549931/0.41578148/0.89616009/1',
 '1/0.1649036/0.30880674/0.93672045/1',
 '1/0.17290651/-0.43181626/0.88523333/1',
 '1/0.17329685/0.18839658/0.96668244/1',
 '1/0.17839998/-0.32509996/0.92869988/1',
 '1/0.17919505/0.05849838/0.98207285/1',
 '1/0.18160174/-0.20480196/0.96180921/1',
 '1/0.18200745/-0.07490306/0.98044012/1',
 '1/0.26939905/0.40849857/0.87209694/1',
 '1/0.28409783/0.30349768/0.90949304/1',
 '1/0.28720483/-0.41910705/0.8613145/1',
 '1/0.29588748/0.18589214/0.93696036/1',
 '1/0.29740177/-0.31410186/0.90160535/1',
 '1/0.30361153/0.05950226/0.95093613/1',
 '1/0.30420604/-0.1965039/0.9321185/1',
 '1/0.30640529/-0.07010121/0.94931639/1',
 '1/0.37379609/0.39649585/0.83849122/1',
 '1/0.39120402/-0.40190413/0.8279085/1',
 '1/0.39179226/0.29439419/0.87168279/1',
 '1/0.40481085/-0.29960803/0.86392315/1',
 '1/0.40607977/0.18089099/0.89575537/1',
 '1/0.41411685/-0.18590756/0.89103626/1',
 '1/0.41498764/0.05959822/0.90787296/1',
 '1/0.41769724/-0.06449957/0.906294/1',
 '1/0.46600106/0.38110087/0.79850181/1',
 '1/0.48278794/-0.38169046/0.78818031/1',
 '1/0.48600297/0.28260173/0.82700506/1',
 '1/0.49848546/-0.28279175/0.8194761/1',
 '1/0.50151639/0.17420569/0.84742769/1',
 '1/0.50918069/-0.1740934/0.84286803/1',
 '1/0.51097962/0.05899765/0.8575658/1',
 '1/0.51360856/-0.05870098/0.85601427/1',
 '1/0.54551481/0.36380988/0.7550205/1',
 '1/0.56150442/-0.35990283/0.74510586/1',
 '1/0.56620006/0.26940003/0.77900008/1',
 '1/0.57807114/-0.26498677/0.77176147/1',
 '1/0.58198189/0.16649482/0.79597523/1',
 '1/0.58933134/-0.1617086/0.7915421/1',
 '1/0.59157958/0.057798/0.80417224/1',
 '1/0.59407609/-0.05289787/0.80266769/1'}{
    
    \shade[ball color = red] (\x*\R,\y*\R,\z*\R) circle (0.08); 
  }
 
\end{tikzpicture}
		}
		\caption{\footnotesize DiLiGenT}
		\label{fig:lamp_diligent}
	\end{subfigure}\qquad
	\begin{subfigure}{0.33\linewidth}
		\resizebox{1.1\linewidth}{!}{
			\tdplotsetmaincoords{180}{180} 
\begin{tikzpicture}[tdplot_main_coords]
\def\R{1.3} 
\def\Raxes{\R*1.3}
\def\angEl{10} 
\foreach \t in {-80,-70,...,80} { \DrawLatitudeCircle[\R]{\t} }
\foreach \t in {5,15,...,175} { \DrawLongitudeCircle[\R]{\t} }

\draw[thick,->] (0,0,0) -- (\Raxes,0,0) node[anchor=north east]{x};
\draw[-stealth] (0,0,0) -- (0,\Raxes,0) node[anchor=north east]{y};
\draw[-stealth] (0,0,0) -- (0,0,\Raxes*1) node[anchor=north east]{z};
\draw[dashed, gray] (0,0,0) -- (-\Raxes,0,0);
\draw[dashed, gray] (0,0,0) -- (0,-\Raxes,0);

\foreach \i/\x/\y/\z/\u in {'1/-0.010081499999999903/0.792314/0.61003/1',
	'1/-0.028800200000000085/-0.688573/0.724595/1',
	'1/-0.05043729999999997/0.228738/0.972181/1',
	'1/-0.05187940000000005/-0.398247/0.91581/1',
	'1/-0.13791699999999993/0.600111/0.787938/1',
	'1/-0.14298500000000003/-0.22794899999999998/0.963117/1',
	'1/-0.15858200000000006/-0.380758/0.910975/1',
	'1/-0.1671780000000001/-0.981945/0.0885159/1',
	'1/-0.2013700000000001/-0.907728/0.368077/1',
	'1/-0.20233799999999988/0.964174/0.171547/1',
	'1/-0.21260699999999988/0.961601/0.173558/1',
	'1/-0.24068099999999992/0.767874/0.593668/1',
	'1/-0.256033/0.08208180000000002/0.963177/1',
	'1/-0.2604649999999999/0.934295/0.243415/1',
	'1/-0.27827799999999997/0.4557510000000001/0.845489/1',
	'1/-0.2947450000000001/-0.816756/0.49602/1',
	'1/-0.2955910000000001/-0.946712/0.127913/1',
	'1/-0.3128950000000001/-0.881633/0.3533/1',
	'1/-0.3225969999999999/0.92739/0.189416/1',
	'1/-0.3241839999999999/0.840594/0.433943/1',
	'1/-0.3298690000000001/-0.570091/0.752451/1',
	'1/-0.37446099999999993/0.23837700000000006/0.896078/1',
	'1/-0.388454/-0.03435059999999995/0.920828/1',
	'1/-0.397339/-0.19131999999999993/0.897507/1',
	'1/-0.42825599999999997/0.594464/0.680595/1',
	'1/-0.46362200000000003/-0.4356499999999999/0.771533/1',
	'1/-0.5075059999999999/0.8546370000000001/0.109696/1',
	'1/-0.5430349999999999/0.6405510000000001/0.542962/1',
	'1/-0.547405/-0.001586909999999933/0.836866/1',
	'1/-0.5559059999999999/0.6899450000000001/0.46362/1',
	'1/-0.5738560000000001/-0.6347469999999998/0.51748/1',
	'1/-0.609997/0.42250400000000005/0.670369/1',
	'1/-0.621389/-0.10455399999999992/0.776495/1',
	'1/-0.6231519999999999/0.7754040000000001/0.102129/1',
	'1/-0.6288519999999999/0.5279790000000001/0.570774/1',
	'1/-0.6443770000000001/-0.7123219999999999/0.278165/1',
	'1/-0.64523/-0.37776099999999996/0.664059/1',
	'1/-0.6475629999999999/0.6821760000000001/0.339556/1',
	'1/-0.6725500000000001/-0.5953979999999999/0.439519/1',
	'1/-0.694801/0.19582000000000008/0.69203/1',
	'1/-0.77267/0.3405120000000001/0.535754/1',
	'1/-0.778546/0.4066700000000001/0.478001/1',
	'1/-0.803476/-0.021562899999999902/0.594947/1',
	'1/-0.8121890000000002/-0.5228029999999999/0.258892/1',
	'1/-0.846169/-0.1606159999999999/0.508134/1',
	'1/-0.849212/-0.4255899999999999/0.312591/1',
	'1/-0.856777/0.14190700000000012/0.495778/1',
	'1/-0.886244/-0.005607909999999891/0.463184/1',
	'1/-0.896355/-0.19160299999999988/0.399795/1',
	'1/-0.93477/-0.2184759999999999/0.280132/1',
	'1/-0.965398/0.03173040000000012/0.258842/1',
	'1/0.0177964999999999/-0.814357/0.580092/1',
	'1/0.03133709999999989/-0.920064/0.390513/1',
	'1/0.06311020000000002/0.133482/0.98904/1',
	'1/0.07259190000000007/0.527221/0.846621/1',
	'1/0.09086539999999996/-0.335585/0.937617/1',
	'1/0.1317920000000001/0.81579/0.563131/1',
	'1/0.14796500000000004/0.16010799999999997/0.975947/1',
	'1/0.17600999999999997/-0.12304400000000003/0.976668/1',
	'1/0.18209199999999992/-0.600153/0.778883/1',
	'1/0.1872580000000001/0.666923/0.721213/1',
	'1/0.2018879999999999/-0.971574/0.123631/1',
	'1/0.24034299999999992/-0.748152/0.618469/1',
	'1/0.26744299999999993/-0.44516000000000006/0.85458/1',
	'1/0.26944900000000005/0.42928099999999997/0.862041/1',
	'1/0.2916919999999999/-0.85004/0.438576/1',
	'1/0.2951740000000001/0.838775/0.457525/1',
	'1/0.30197900000000005/0.566708/0.766584/1',
	'1/0.3573080000000001/0.926392/0.118866/1',
	'1/0.397158/0.09466119999999995/0.912855/1',
	'1/0.4088689999999999/-0.28941800000000006/0.865485/1',
	'1/0.41418300000000013/0.81302/0.409208/1',
	'1/0.45522500000000005/0.33174599999999993/0.826266/1',
	'1/0.4686210000000001/0.8736879999999999/0.130627/1',
	'1/0.47120399999999996/-0.5565790000000002/0.684242/1',
	'1/0.4727499999999999/-0.7481750000000001/0.465555/1',
	'1/0.4734879999999999/-0.8615940000000001/0.182934/1',
	'1/0.4953890000000001/0.44450599999999996/0.746327/1',
	'1/0.5413379999999999/-0.5680010000000001/0.619942/1',
	'1/0.555808/-0.006201030000000067/0.831288/1',
	'1/0.5978460000000001/0.6118269999999999/0.517926/1',
	'1/0.605934/-0.2484850000000001/0.755711/1',
	'1/0.6273709999999999/-0.5751580000000001/0.524975/1',
	'1/0.663553/-0.14739400000000008/0.733466/1',
	'1/0.665535/0.1987619999999999/0.719414/1',
	'1/0.6900449999999999/-0.49172500000000013/0.531079/1',
	'1/0.694354/0.3760039999999999/0.61359/1',
	'1/0.6974680000000001/0.7063839999999999/0.120668/1',
	'1/0.7059020000000001/0.5043229999999999/0.497353/1',
	'1/0.7407709999999998/-0.6595290000000001/0.127595/1',
	'1/0.795127/-0.31366500000000014/0.519026/1',
	'1/0.816713/0.1605539999999999/0.554258/1',
	'1/0.819005/-0.08370620000000009/0.567648/1',
	'1/0.837542/0.04648879999999989/0.544391/1',
	'1/0.8735219999999999/-0.4711490000000001/0.122381/1',
	'1/0.892605/0.3690789999999999/0.258914/1',
	'1/0.894863/0.3635919999999999/0.258883/1',
	'1/0.900714/-0.3809470000000001/0.208791/1',
	'1/0.943202/0.19742699999999988/0.267193/1',
	'1/0.955201/-0.2428290000000001/0.16919/1'}{
    
    \shade[ball color = red] (\x*\R,\y*\R,\z*\R) circle (0.08); 
  }
 
\end{tikzpicture}
		}
		\caption{\footnotesize DiLiGenT10$^2$}
		\label{fig:lamp_diligent_10}
	\end{subfigure}\\[.5em]
	\begin{subfigure}{0.33\linewidth}
		\resizebox{1.1\linewidth}{!}{
			\tdplotsetmaincoords{180}{180} 
\begin{tikzpicture}[tdplot_main_coords]
\def\R{1.3} 
\def\Raxes{\R*1.3}
\def\angEl{10} 
\foreach \t in {-80,-70,...,80} { \DrawLatitudeCircle[\R]{\t} }
\foreach \t in {5,15,...,175} { \DrawLongitudeCircle[\R]{\t} }

\draw[thick,->] (0,0,0) -- (\Raxes,0,0) node[anchor=north east]{x};
\draw[-stealth] (0,0,0) -- (0,\Raxes,0) node[anchor=north east]{y};
\draw[-stealth] (0,0,0) -- (0,0,\Raxes*1) node[anchor=north east]{z};
\draw[dashed, gray] (0,0,0) -- (-\Raxes,0,0);
\draw[dashed, gray] (0,0,0) -- (0,-\Raxes,0);

\foreach \i/\x/\y/\z/\u in {'1/-0.0089/0.056/0.9984/1',
 '1/-0.0554/0.2725/0.9606/1',
 '1/-0.0563/-0.0882/0.9945/1',
 '1/-0.0893/0.1153/0.9893/1',
 '1/-0.0992/0.0034/0.9951/1',
 '1/-0.3016/-0.8628/0.4056/1',
 '1/-0.3423/0.5797/0.7394/1',
 '1/-0.3505/-0.3641/0.8629/1',
 '1/-0.4388/0.2469/0.864/1',
 '1/-0.4841/-0.0775/0.8716/1',
 '1/-0.4962/0.7352/0.4619/1',
 '1/-0.5471/-0.1063/0.8303/1',
 '1/-0.567/-0.2087/0.7969/1',
 '1/-0.644/-0.606/0.4669/1',
 '1/-0.6519/0.2109/0.7284/1',
 '1/-0.6857/-0.7023/0.1911/1',
 '1/-0.7165/-0.0295/0.6969/1',
 '1/-0.7925/0.4995/0.3498/1',
 '1/-0.8089/-0.2354/0.5388/1',
 '1/-0.9107/-0.2802/0.3036/1',
 '1/-0.9251/0.2552/0.2811/1',
 '1/-0.9584/0.2415/0.1521/1',
 '1/-0.9723/-0.1494/0.1795/1',
 '1/0.0129/0.4477/0.8941/1',
 '1/0.0547/0.0041/0.9985/1',
 '1/0.0726/-0.7866/0.6131/1',
 '1/0.097/-0.7923/0.6024/1',
 '1/0.098/0.5064/0.8567/1',
 '1/0.1927/0.2997/0.9344/1',
 '1/0.2105/-0.6831/0.6993/1',
 '1/0.2308/-0.1355/0.9635/1',
 '1/0.2404/0.191/0.9517/1',
 '1/0.2622/-0.7088/0.6548/1',
 '1/0.2799/-0.0131/0.9599/1',
 '1/0.287/-0.5394/0.7916/1',
 '1/0.2938/-0.9199/0.2597/1',
 '1/0.2992/-0.5057/0.8092/1',
 '1/0.3227/0.8283/0.458/1',
 '1/0.3406/0.0568/0.9385/1',
 '1/0.3691/-0.6913/0.6211/1',
 '1/0.4214/-0.5787/0.6982/1',
 '1/0.4282/-0.5824/0.691/1',
 '1/0.4695/-0.4364/0.7675/1',
 '1/0.5072/0.7878/0.3495/1',
 '1/0.5235/-0.4552/0.7203/1',
 '1/0.5369/-0.5873/0.6057/1',
 '1/0.5695/0.6038/0.5578/1',
 '1/0.5888/-0.4187/0.6914/1',
 '1/0.6136/0.4207/0.6682/1',
 '1/0.616/0.1628/0.7708/1',
 '1/0.6602/-0.412/0.628/1',
 '1/0.6617/0.2285/0.7141/1',
 '1/0.6836/-0.6105/0.4/1',
 '1/0.6942/-0.2329/0.681/1',
 '1/0.715/0.1689/0.6784/1',
 '1/0.7181/0.5923/0.3653/1',
 '1/0.7547/0.638/0.1527/1',
 '1/0.8755/0.4821/0.0313/1',
 '1/0.8939/-0.1615/0.4181/1',
 '1/0.897/0.282/0.3403/1',
 '1/0.8975/0.3988/0.1885/1',
 '1/0.938/0.2845/0.1979/1',
 '1/0.9511/-0.0526/0.3042/1',
 '1/0.9632/-0.2114/0.166/1'}{
    
    \shade[ball color = red] (\x*\R,\y*\R,\z*\R) circle (0.08); 
  }
 
\end{tikzpicture}
		}
		\caption{\footnotesize Random}
		\label{fig:lamp_random}
	\end{subfigure}
		\vspace*{.75em}
	\caption{Distribution of illumination directions in the real \emph{DiLiGenT} and \emph{DiLiGenT10}$^2$ datasets, and an example of a random distribution. The $z$-axis corresponds to the optical axis of the camera, with the imaged object at coordinates $(0,0,0)$.\vspace*{-.25em}}
	\label{fig:lamp_position}
\end{figure}
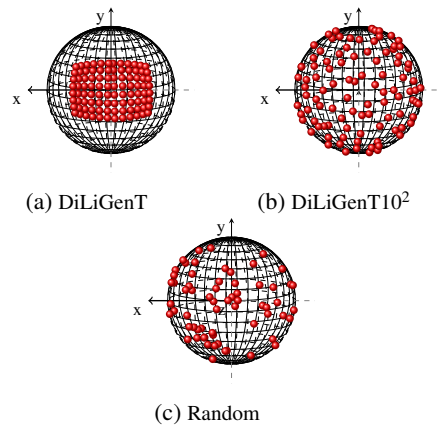

The \emph{Blobby} dataset is composed of $10$ geometric shapes, each one observed from 1296 distinct viewpoints. As the name suggests, the shapes in \emph{Blobby} are rather smooth and regular (Fig.~\ref{fig:datasets}a). The \emph{Structure} dataset consists in objects with complex geometry containing fine details (Fig.~\ref{fig:datasets}b). It is composed of $8$ objects, rendered in 3D from $1387$ to $6874$ viewpoints. To simulate surfaces with non-Lambertian light reflectance, a material from the \emph{MERL}~\cite{MERL} dataset is randomly drawn and applied in each rendering, providing a total of $25920$ samples for \emph{Blobby} and $59292$ for \emph{Structure}. In both cases, each sample is rendered under $64$ different light directions, randomly selected on the hemisphere (Fig.~\ref{fig:lamp_random}). 

Finally, the \emph{CyclePS}~\cite{CNN_PS} dataset is also composed of complex objects, but contains only 18 objects rendered from 10 views (Fig.~\ref{fig:datasets}c). However, the number of materials available is substantial because \emph{Disney's principled BSDF}~\cite{burley2012physically} parametric reflectance model is used. It allows the variation of the base colour, roughness, proportion of specular reflectance, etc., thus the objects can be rendered using a near infinite number of materials. The training dataset presented in the present paper will also feature the possibility to generate as many materials as needed, while also considering much more geometric shapes than in existing sets.

\paragraph{Existing benchmarking datasets}

To validate the relevance of the training datasets, as well as to verify that the models trained on these synthetic data are able to generalize to real images, two real-world datasets exist: \emph{DiLiGenT}~\cite{dilidataset} and $\emph{DiLiGenT10}^2$~\cite{DILI_10}. 

The \emph{DiLiGenT} dataset comprises 10 different objects, taken from the same viewpoint under $96$ different illuminations (Fig.~\ref{fig:lamp_diligent}). The reflectance of the objects in this dataset goes from quasi-Lambertian to moderately specular. For each photographed object, the ground truth normal map is provided, as well as the calibrated lighting directions and intensities. Therein, the ground truth geometry was acquired by manually registering laser scans with the images. 

The $\emph{DiLiGenT10}^2$ dataset contains 10 different objects. Each object was explicitly fabricated with 10 different materials and photographed under 100 calibrated illuminations (Fig.~\ref{fig:lamp_diligent_10}). The ground truth was not obtained by scanning the objects, but from the 3D digital models used to machine the objects. This real dataset is particularly interesting for evaluating performances on highly specular materials and translucent ones. Indeed, it contains metallic materials, such as aluminium or steel, and a translucent one (acrylic). This dataset also contains diffuse and slightly specular materials, hence most of real-world material characteristics are present. The diversity of object shapes is also high as it contains objects with simple geometry like balls but also complex ones like turbines. It offers the opportunity to test the impact of diverse inter-reflection, shadow and shading effects. Today, it is the most complete dataset composed of \emph{real} images available in PS.

\paragraph{Uncalibrated PS}

In all the methods discussed above, the light directions and intensities are assumed to be known, i.e.\ we consider the \emph{calibrated} PS problem. When these acquisition parameters are unknown, the problem is called \emph{uncalibrated}. Uncalibrated PS has been studied e.g.\ in~\cite{SDPS_net,uni_ps,uncalibrated_neural_inverse}, and partially solved by defining a first neural network that predicts the lighting parameters associated with each acquired image.
This estimated data is then fed into a second network that solves the problem of calibrated PS. Managing non-directional lighting, e.g.\ near point-light sources~\cite{logothetis2022cnn,santo2020deep} or natural illumination~\cite{haefner2019variational,uni_ps,mo2018uncalibrated}, is another ongoing research problem. In this paper we focus on the case of \emph{calibrated} PS with known \textit{directional} light sources.

\section{A New Multi-Scale Architecture for PS}
\label{sec:network_archi}

\noindent The multi-scale architecture we propose builds upon the normal estimation network introduced in \cite{PS_FCN}. Therein, each image is first normalized by the calibrated lighting intensity, and then concatenated with the calibrated direction. The resulting ``image'' forms the input to the feature extractor which processes each (image, direction) pair independently. Then, all the independent features are aggregated through a feature aggregation module, and lastly a regression module predicts the normal map.

\begin{figure}[!htpb]
	\centering
	\subfloat{\includegraphics[width=0.312\linewidth]{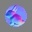}}\hspace{0.05cm}
	\subfloat{\includegraphics[width=0.312\linewidth]{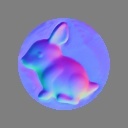}}\hspace{0.05cm}
	\subfloat{\includegraphics[width=0.312\linewidth]{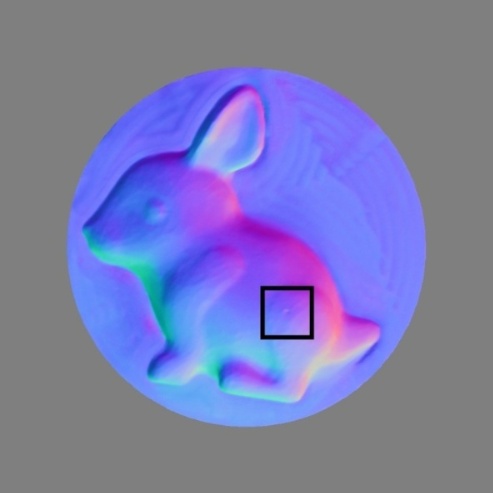}}\hspace{0.05cm}
	\\
	\vspace{0.05cm}
	\hspace{0.101cm}\subfloat{\includegraphics[width=0.312\linewidth]{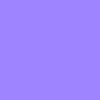}}\hspace{0.05cm}
	\subfloat{\includegraphics[width=0.312\linewidth]{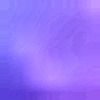}}\hspace{0.05cm}
	\subfloat{\includegraphics[width=0.312\linewidth]{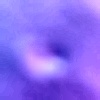}}\hspace{0.05cm}
	\vspace*{0.5em}
	\caption{Multi-scale normal estimation at three different scales (bottom row is a contrast-enhanced zoom on the rectangle area). Low-detail geometry is reconstructed from the first levels. High-frequency details get refined as the scale increases.}
	\label{fig:detail_multi}
\end{figure}

In order for the normal estimation to perform equivalently well on low-frequency geometry and high-frequency details, we propose to embed this network in a \emph{multi-scale} approach which progressively refines the result as the spatial scale increases. Thus, our model first focuses on the \emph{global} aspect of the object, then progressively insert \emph{details} such as cracks, slight bumps, or holes as illustrated in Fig.~\ref{fig:detail_multi}.

\begin{figure*}[!htpb]
	\centering
	\includegraphics[width=\linewidth]{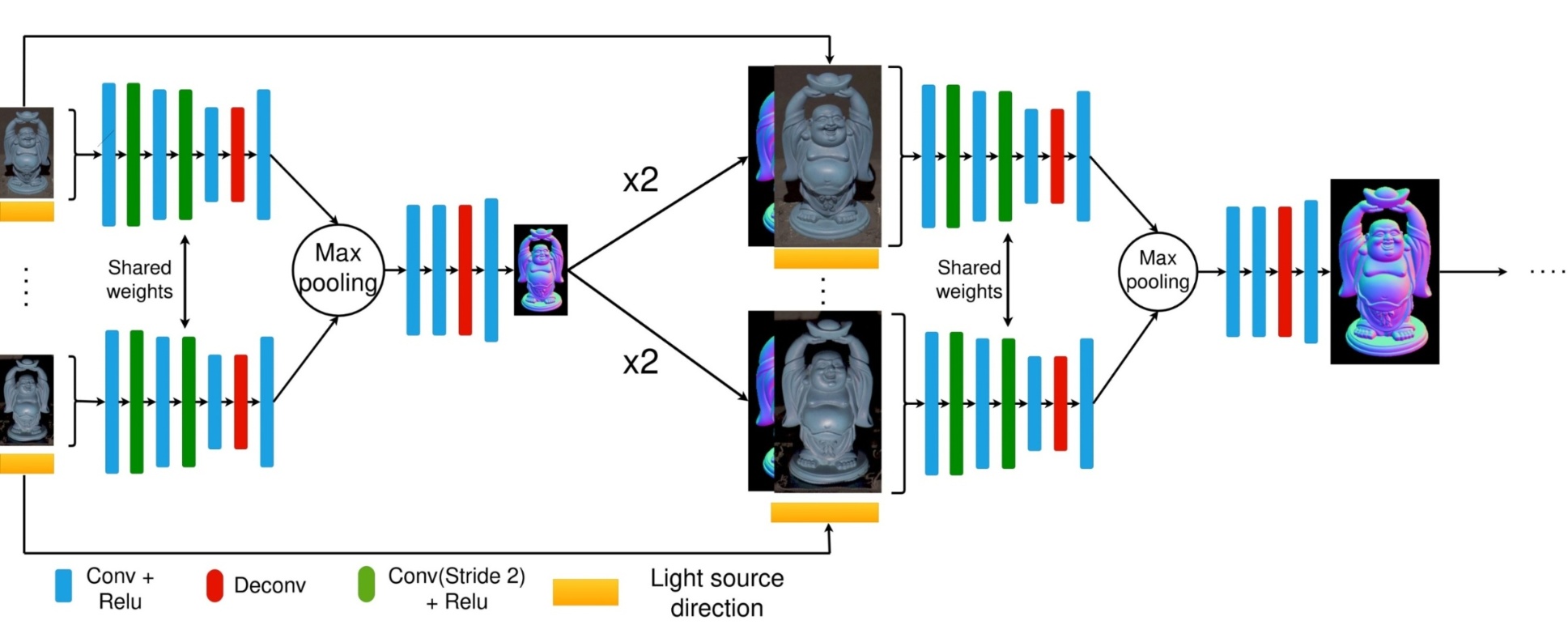}
	\caption{First two stages of the proposed multi-scale architecture. A first architecture, inspired by the PS-FCN method~\cite{PS_FCN}, takes as inputs the calibrated lighting directions and downsampled images, and outputs a low-resolution 3D normal map. The latter is then up-sampled and concatenated with lighting directions and higher-resolution images. A second architecture then infers higher-resolution normals, and this part of the process is repeated until the resolution of the original images is reached (network weights being shared by all scales).}
	\label{fig:multi_scale_archi}
	\vspace*{0.5em}
\end{figure*}

The proposed multi-scale network combines two independent architectures (Fig.~\ref{fig:multi_scale_archi}). The first stage takes as inputs the calibrated lighting directions and the images (downsampled from the original images to some initial resolution $r_0$), and outputs a low-resolution normal map with the same resolution $r_0$. This first stage is essentially similar to the normal estimation network proposed in~\cite{PS_FCN}. In the second stage, the low-resolution 3D normal map is up-sampled to a resolution $r_1 = 2 r_0$ (using bilinear interpolation followed by normalization to enforce the unit-length constraint on normal vectors), and concatenated with the images (down-sampled from the original input images to resolution $r_1$) and lighting directions. The process is then repeated until the resolution of the original images is reached. In these sequential stages, the inputs differ from the first stage, thus a new, independent architecture is obviously necessary. Yet, let us emphasize that since this new architecture is completely convolutional (except the pooling layer) and as only the spatial resolution changes from stage to stage, we can share the weights between each processed scale. Therefore, only two networks actually need being trained, independently from the number of scales. The network formed by these two sub-networks is trained by minimizing the cosine similarity, which measures the angular difference between the estimated 3D normals and the ground truth ones. It is defined as follows:
\begin{equation}
l_{normal} = 1 - \sum_{ij} N_{ij}^\top \hat{N}_{ij},
\end{equation}
where $\hat{N}_{ij}$ is the estimated normal at pixel $(i,j)$, and $N_{ij}$ is the ground truth one.
In terms of computational cost, our multi-scale CNN has 4.4 millions parameters. In comparison with the mono-scale approach, it uses only 5\% more memory and takes 14\% more time for inference.

As remarked in~\cite{shape_and_material}, one of the most interesting features of a multi-scale architecture is its ability to process images with arbitrary size (small or large) without loss of performance. Indeed, even if a single-scale model is fully convolutional and so can process high-resolution images, such a model with a fixed number of convolution layers may not have enough convolutions to synthesize the information over a whole, potentially large image. And, a network trained to handle a specific resolution may not behave well for much larger images. For example, information from the bottom left of the image may not be used to infer the normal at the top right. Yet, such an ability would be particularly useful for handling non-local reflectance effects such as translucency. See for instance the acrylic ball shown in the experiments section in Fig.~\ref{fig:zoom_acrylique_translucide}, where light passes through the object.
By propagating global information at different scales, such a limitation of local methods is overcome.

More importantly, the proposed multi-scale architecture with shared weights allows one to process images with higher resolution than the ones used during training.  For example, in our implementation the first processing resolution is $8\times 8$ pixels. By taking a resolution multiplier of two between two scales, four scales are necessary to reach a resolution of $128\times 128$ pixels (which is the training resolution in our tests), and seven scales for the $\emph{DiLiGenT10}^2$ images which have a resolution of $1001\times 1001$ pixels. Yet, the same weights are used in both cases, hence a resolution-specific training is not necessary. In practice, this removes the need for either rescaling the input images to the resolution of the training images, or resorting to a (too local) patch-based approach.

\section{Proposed Learning Dataset}
\label{sec:proposed_dataset}

As discussed in Section~\ref{sec:relatedworks}, the existing \emph{Blobby} and \emph{Structure} synthetic datasets lack of diversity in terms of geometry and textures. For example, although the \emph{Structure} dataset is composed of complex objects, all these objects are statues.
Similarly, the number of different materials in the MERL material base is only 100. This is clearly not enough to model the huge diversity of materials present in the nature. The \textit{CyclePS} dataset partially solves this issue, by allowing to generate infinitely many materials by randomly selecting parameters from a parametric BSDF model. Still, it remains limited in terms of geometry. Overall, a greater diversity of shapes and materials in the images of the training dataset would be beneficial for training networks for photometric stereo. For these reasons, we propose here a new dataset, which includes a large variety of shapes and materials.

\begin{table*}[h]
	\footnotesize
	\resizebox{0.99\linewidth}{!}{
		\begin{tabular}{p{3cm}|c|c|c|c|c}
			& $\#$ objects & $\#$ views & $\#$ total number of samples & $\#$ lighting & $\#$ materials \\
			\hline
			\textit{Blobby} & 10 & 1 296 & 25 920 & 64 & 100 \\
			\textit{Structure} & 8 & 1387-6874 & 59 292 & 64 & 100 \\
			\emph{CyclePS} & 18 & 10 & 180 & 1 300 & 90 000\\
			\emph{DiLiGenT} & 10 & 1 & 10 & 96 & 10 \\
			\emph{DiLiGenT10$^2$} & 10 & 10 & 100 & 100 & 10 \\
			\emph{Our Blobby} & 3000 &  5 & 15 000 & 100 & 1 100 + infinity \\
			\emph{Our Object} & 76 & 267 & 45 000 & 100 & 1 100 + infinity \\
		\end{tabular}
	}\\[.5em]
	\caption{Summary of the characteristics of the different learning datasets used in photometric stereo.}
	\label{Tab:tableau_donnees}
\end{table*}

\begin{figure}[!ht]
	\centering
	\subfloat{\includegraphics[width=0.321\linewidth]{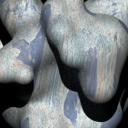}}\hspace{0.05cm}
	\subfloat{\includegraphics[width=0.321\linewidth]{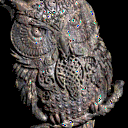}}\hspace{0.05cm}
	\subfloat{\includegraphics[width=0.321\linewidth]{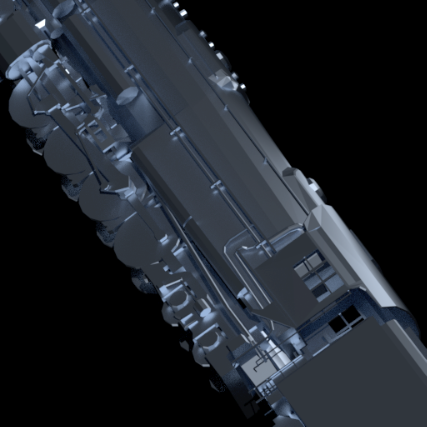}}\\
	\vspace{0.05cm}
	\subfloat{\includegraphics[width=0.321\linewidth]{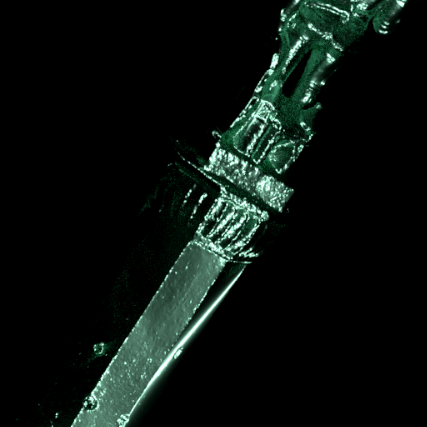}}\hspace{0.05cm}
	\subfloat{\includegraphics[width=0.321\linewidth]{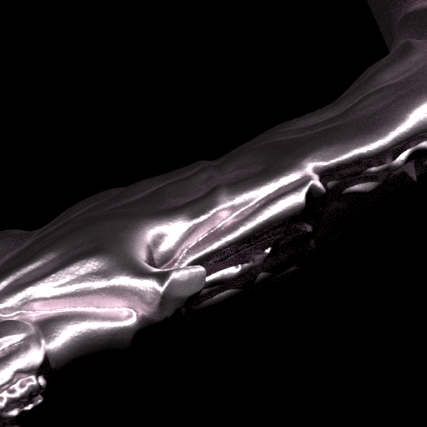}}\hspace{0.05cm}
	\subfloat{\includegraphics[width=0.321\linewidth]{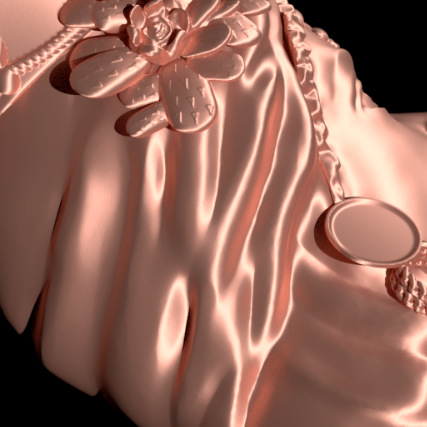}}
	\vspace*{0.8em}
	\caption{Examples of images from the proposed dataset.\vspace*{-1.5em}}
	\label{fig:datasets_GREYC}
\end{figure}

In order to create this dataset, we implemented our own image data generation pipeline. We used the \emph{Blender}~\cite{blender} software with the Cycles rendering engine. As a result, our new dataset is composed of two parts:
\begin{itemize}
	\item \textit{Our Blobby} contains objects with smooth surfaces;
	\item \textit{Our Object} contains objects with complex geometry: strong discontinuities, edges, corners, textures details, etc.
\end{itemize}
Samples from our training dataset are shown in Fig.~\ref{fig:datasets_GREYC}.

\noindent\textit{Our Blobby} has 3000 distinct objects, generated by the sum of random Gaussian potentials, followed by iso-surface extraction using the \emph{Marching Cubes} algorithm~\cite{marching}. \textit{Our Object} contains 76 detailed objects which are 3D meshes from the \emph{Sketchfab}~\cite{Sketchfab} website. Moreover, to allow the learning of non-Lambertian surfaces, more than 1100 different real materials, extracted from the \emph{ambientCG}~\cite{Ambientcg} website, are randomly applied to the objects, much more than the 100 materials of \emph{Structure} and \emph{Blobby}. To complete a lack of diversity of the most complicated materials (metals, glasses, etc.) that could persist, we generated additional materials by randomly setting the values of somes parameters (metallic, specular, roughness, anisotropic, etc.) of Disney's principled BSDF~\cite{burley2012physically}. To ensure that all possible materials are represented, during the rendering we choose~to~apply to the object with a probabilty of 50\% a real material (from ambientCG), with 17\% a glass material and with 17\% a metal one. The remaining $16\%$ materials are constructed by randomly selecting all possible parameters in the principled BSDF (which may result in non-realistic materials).

\begin{figure}[!ht]
	\centering
	\captionsetup[subfigure]{labelformat=empty}
	\subfloat[Uniform reflectance]{\label{fig:render_uniform}\includegraphics[width=0.375\linewidth]{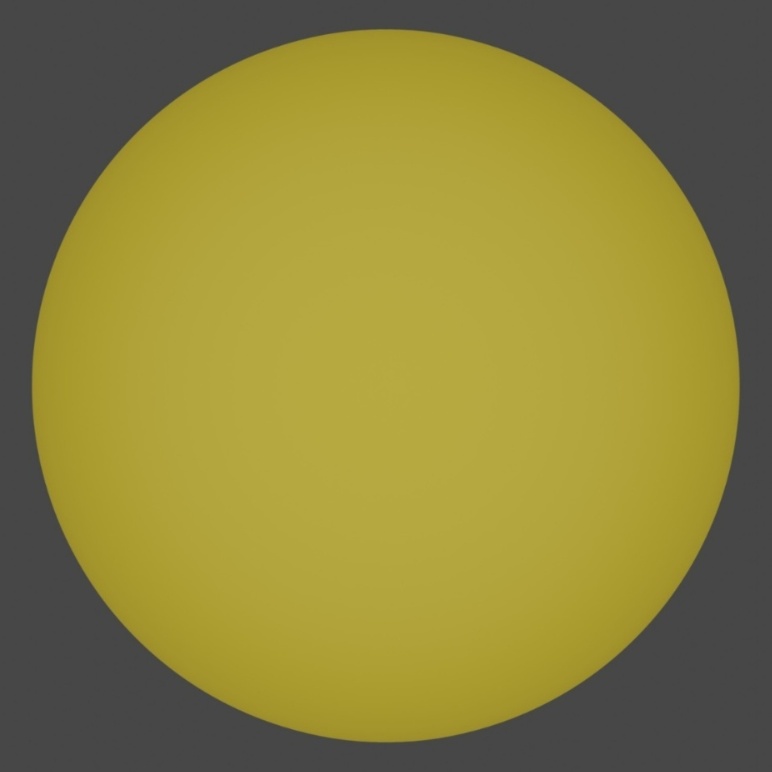}}\hspace{0.05cm} \qquad
	\subfloat[\centering Spatially-varying reflectance]{\label{fig:render_variation}\includegraphics[width=0.375\linewidth]{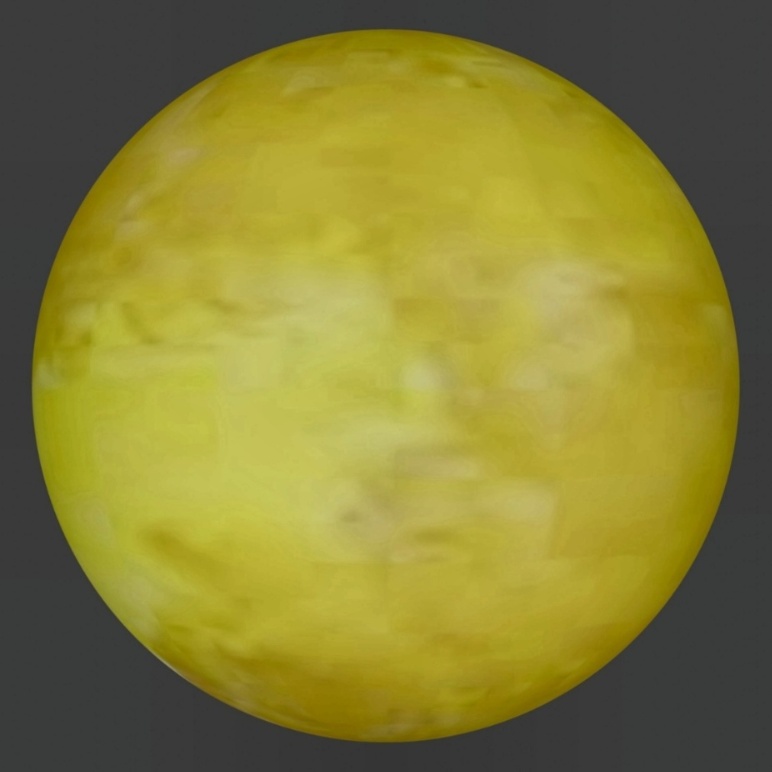}}
	\vspace*{0.5em}
	\caption{Rendering of the same ball with a uniform base color, or with a spatially-varying one.}
	\label{fig:comparison_uniform_render}
\end{figure}

If we set a single value for each parameter of the principled BSDF, we would obtain a material which is spatially uniform in terms of reflectance, as in the example of Fig.~\ref{fig:render_uniform}. Yet, many real-world objects exhibit a spatially-varying reflectance, which is a known limitation of existing PS techniques~\cite{PS_FCN}. To solve this problem in our generation pipeline, we rather incorporated a few spatially-varying material maps, as in the example of Fig.~\ref{fig:render_variation}. This technique was used for 50\% of the renderings. It allowed us to create both objects with uniform reflectance, and others with spatially-varying one, as illustrated in Fig.~\ref{fig:datasets_GREYC}.

Finally, to generate data having realistic lighting conditions, we rendered all the images with both random illumination direction (Fig.\ref{fig:lamp_random}) and random intensity. In total, 15 000 \textit{blobby} samples and 45 000 \textit{object} samples were generated this way. Table~\ref{Tab:tableau_donnees} summarizes the characteristics of the existing datasets, versus the ones we propose. In order to ensure the reproducibility of our results, the code and these learning datasets will be made publicly available online.

\section{Experiments}

In this section, we demonstrate the effectiveness of our proposed multi-scale architecture on publicly available benchmarks, namely DiLiGenT~\cite{dilidataset} and DiLiGenT10$^2$~\cite{DILI_10}. To evaluate the impact of our new training dataset, we trained our network both on the pre-existing training datasets \textit{Blobby} and \textit{Structure} (this training is referred to as ``DS1'' in the following) and on our new training dataset (``DS2'' in the following). In the rest of this section, ``\emph{Mono} (DS1)'' will thus refer to the mono-scale architecture trained on the pre-existing dataset, ``\emph{Multi} (DS1 + DS2)'' to the multi-scale architecture trained on both the pre-existing and the new datasets, etc. We will first provide a few qualitative results to illustrate the importance of the two building blocks of our contribution, and then provide a thorough quantitative evaluation on the two benchmarks.

\subsection{Implementation details} Both the ``\emph{Mono}'' and the ``\emph{Multi} architectures were implemented in Pytorch. The Adam optimizer~\cite{adam} was used with a learning rate of $10^{-4}$. We trained both the multi-scale and the mono scale architecture by taking 32 patches of size 128 by 128 as inputs. The training took a few days on a single Nvidia GeForce GTX 1080 Ti with a batch size of 3 (the maximun we can fit in our GPU). The inference time depends on the number of input images and their resolution. For example, by taking 100 images of 256 by 256 pixels, it takes approximately 1.6 seconds for our multi-scale methods on our GPU. The inference time scales linearly with the number of images, while it seems to be roughly multiplied by a factor of 4 when the resolution is multiplied by 2.

\subsection{Qualitative evaluation}

\begin{figure}[!ht]
	\centering
	\captionsetup[subfigure]{labelformat=empty}
	\subfloat{\includegraphics[width=0.321\linewidth]{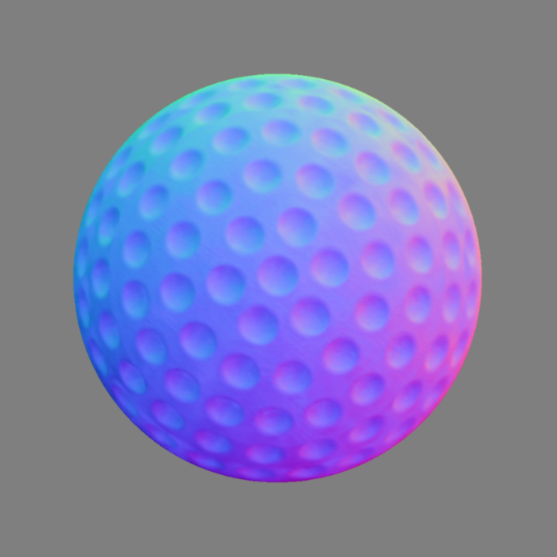}}\hspace{0.05cm}
	\subfloat{\includegraphics[width=0.321\linewidth]{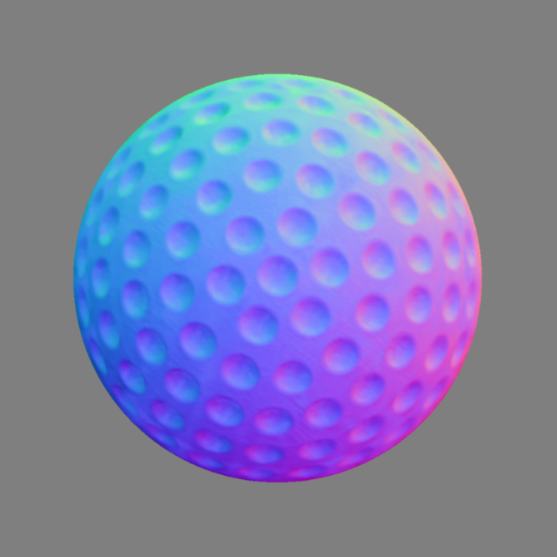}}\hspace{0.05cm}
	\subfloat{\includegraphics[width=0.321\linewidth]{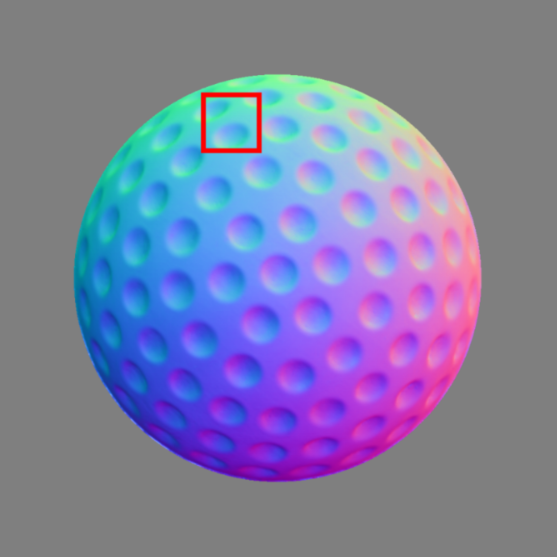}}\hspace{0.05cm}	\\[.5em]
	\subfloat[\emph{Mono} (DS1)]{\includegraphics[width=0.321\linewidth]{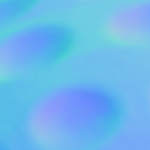}}\hspace{0.05cm}
	\subfloat[\emph{Multi} (DS1)]{\includegraphics[width=0.321\linewidth]{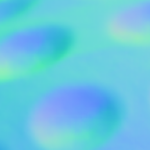}}\hspace{0.05cm}
	\subfloat[``Ground truth'']{\includegraphics[width=0.321\linewidth]{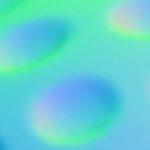}}\hspace{0.05cm} \vspace*{0.8em}
	\caption{Results of our mono- and multi-scale architectures (both trained on the pre-existing dataset DS1) on the copper golf ball from~\cite{DILI_10}. The multi-scale architecture yields much sharper results, especially around the holes.}
	\label{fig:mono_vs_multi}
\end{figure}

Let us start by showing two illustrative results on the DiLiGenT10$^2$~\cite{DILI_10} benchmark, on challenging metallic objects (the copper golf ball and the copper hexagon). As we shall see, both the new training dataset and the new multi-scale architecture contribute to improving the estimation performances on such objects exhibiting an anisotropic reflectance. Since we do not have access to the ground truth normals, for visual purpose we show as ``ground truth'' the result we obtained with our \emph{Multi} (DS1+DS2) approach, applied to the same object but fabricated in PVC (a matte material). The example of Fig.~\ref{fig:mono_vs_multi} shows that, independently from the training set, the multi-scale architecture largely contributes to improving the results on metals. In this example, the same dataset is used for training both the mono-scale and multi-scale architectures, and the latter offers visually more accuate results. Likely, the ability of the multi-scale architecture to propagate information in a global manner helps interpreting the anisotropic behavior. 

\begin{figure}[!ht]
	\centering
	\captionsetup[subfigure]{labelformat=empty}
	\subfloat[\emph{Mono} (DS1)]{\includegraphics[width=0.321\linewidth]{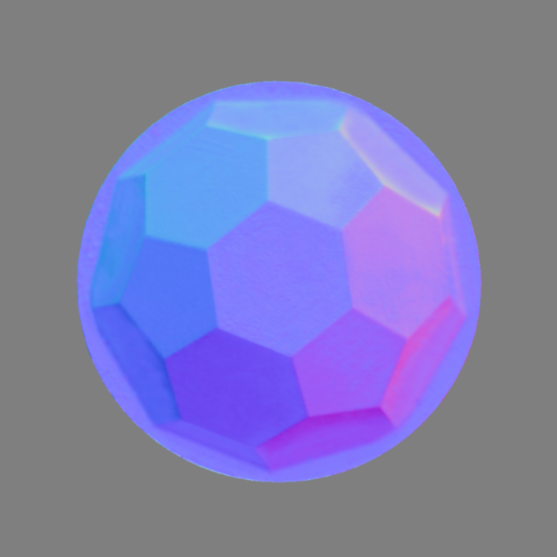}}\hspace{0.05cm}
	\subfloat[\emph{Mono} (DS1+DS2)]{\includegraphics[width=0.321\linewidth]{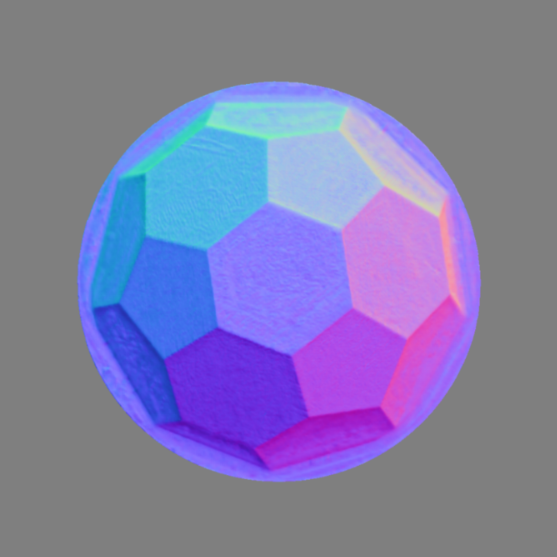}}\hspace{0.05cm}
	\subfloat[``Ground truth'']{\includegraphics[width=0.321\linewidth]{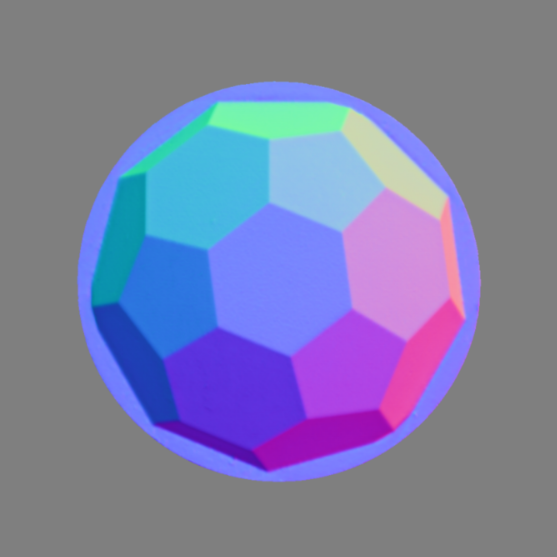}}\hspace{0.05cm}
	\vspace*{0.8em}
	\caption{Results of our mono-scale architecture on the copper hexagon from~\cite{DILI_10}. Since the new dataset (DS2) contains much more metallic objects than the existing one (DS1), training on our new dataset yields largely improved results.}
	\label{fig:mono_old_new}
\end{figure}

The example of Fig.~\ref{fig:mono_old_new}, on the contrary, shows the importance of the presence of metallic objects in the training dataset, independently from the network architecture. It can be observed that the network performs much better when it is trained on our new training dataset, even without considering the multi-scale architecture.

\begin{figure}[!ht]
	\centering
	\captionsetup[subfigure]{labelformat=empty}

	\begin{minipage}{0.22\linewidth}
		\subfloat[Acrylic ball]{\includegraphics[width=\linewidth]{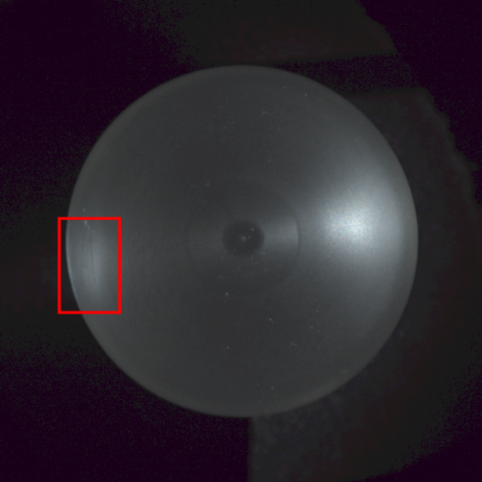}}
	\end{minipage}
	\begin{minipage}{0.75\linewidth}
		\subfloat{\includegraphics[width=0.315\linewidth]{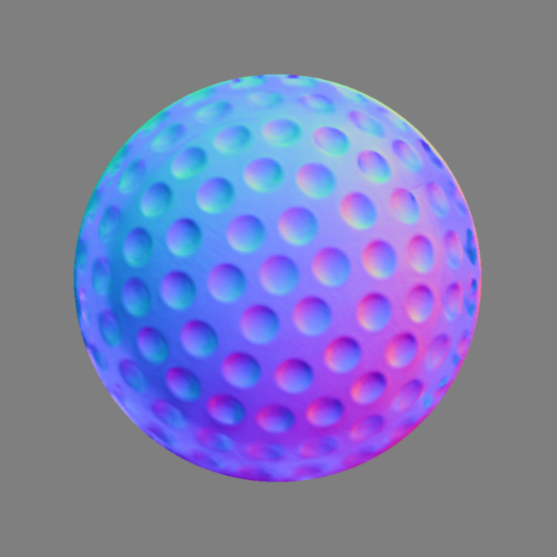}}\hspace{0.05cm}
		\subfloat{\includegraphics[width=0.315\linewidth]{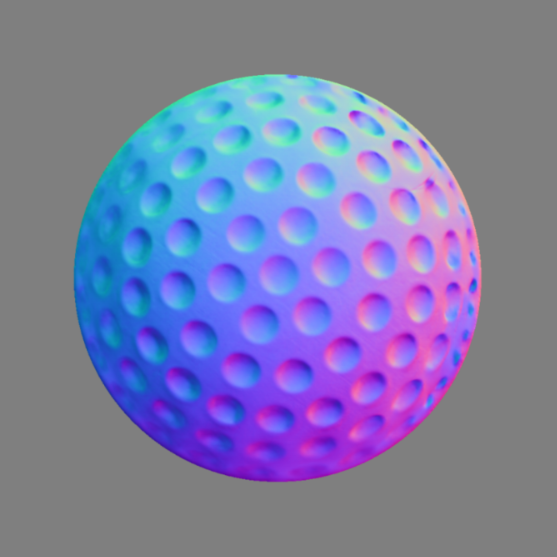}}\hspace{0.05cm}
		\subfloat{\includegraphics[width=0.315\linewidth]{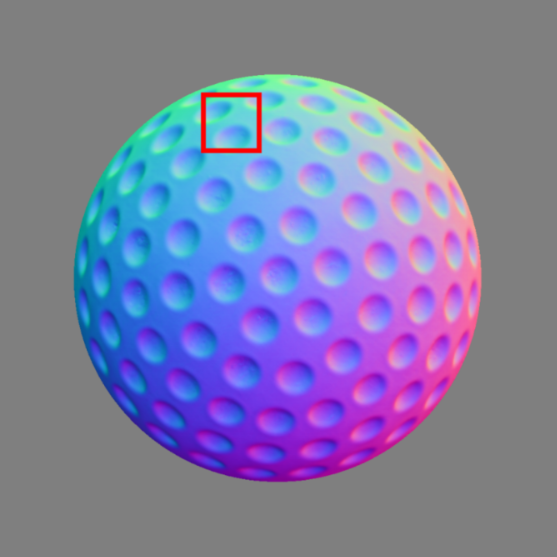}}
		
		\vspace*{0.1em}
		
		\subfloat[\centering \emph{Mono} (DS1+DS2)]{\includegraphics[width=0.315\linewidth]{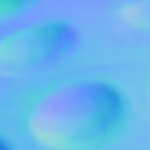}}\hspace{0.05cm}
		\subfloat[\centering \emph{Multi} (DS1+DS2)]{\includegraphics[width=0.315\linewidth]{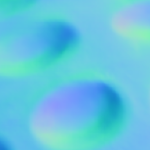}}\hspace{0.05cm}
		\subfloat[``Ground truth'']{\includegraphics[width=0.315\linewidth]{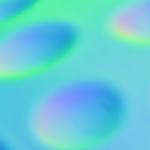}}
		
	\end{minipage}\vspace*{0.8em}
	\caption{An image of an acrylic ball from~\cite{DILI_10}, illuminated from the right, and results of our mono- and multi-scale architectures (both trained on the new dataset DS2) on the acrylic golf ball from~\cite{DILI_10}. The reconstruction of translucent objects is improved a lot by using the multi-scale approach.}
	\label{fig:zoom_acrylique_translucide}
\end{figure}

Fig.~\ref{fig:zoom_acrylique_translucide} illustrates a particularly visible improvement brought by the multi-scale architecture, which is the correct handling of translucent materials. In this example, we consider again the gulf ball from~\cite{DILI_10}, but this time coated with an acrylic material. Acrylic is a glass-like material, with some of the light passing through the object. As can be seen in the top of Fig.~\ref{fig:zoom_acrylique_translucide}, even when light comes from the right side of the ball, part of its left side appears illuminated. Without seeing the whole object the model could not imagine that there exists a path underneath the surface that lets the light go through. On the contrary, the multi-scale approach being global by construction, such non-local phenomena are better managed by the network and the overall reconstruction is clearly more accurate.

\begin{figure}[H]
	\captionsetup[subfigure]{labelformat=empty}
	\begin{minipage}{0.1\linewidth}
		\subfloat[]{\includegraphics[height=3.5\linewidth]{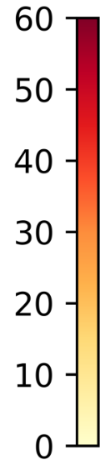}}
	\end{minipage}
	\begin{minipage}{0.9\linewidth}
		\subfloat[]{\includegraphics[height=0.335\linewidth]{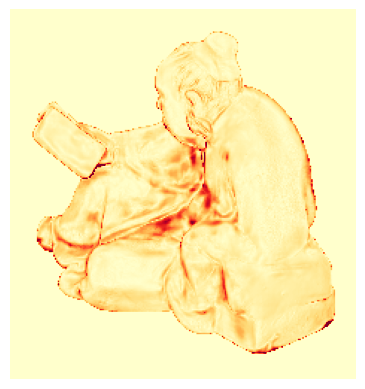}}
		\hspace{-0.15cm}
		\subfloat[]{\includegraphics[height=0.335\linewidth]{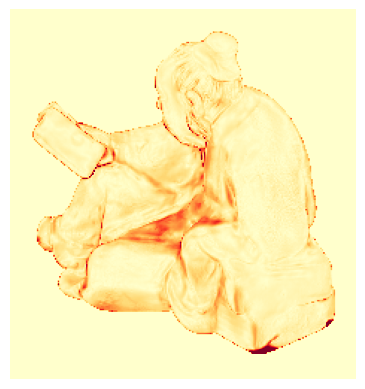}}
		\hspace{-0.06cm}
		\subfloat[]{\includegraphics[width=0.305\linewidth]{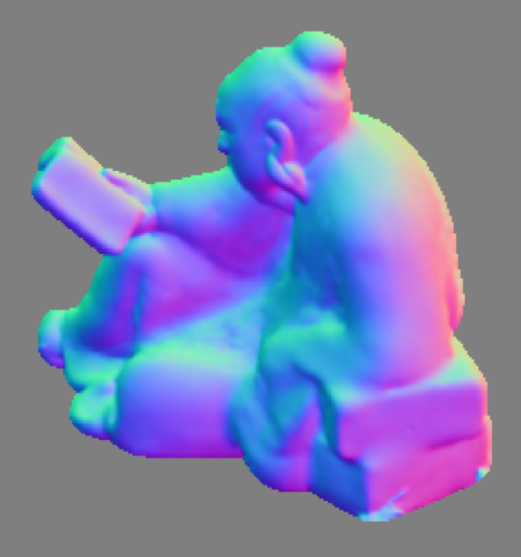}}
		\vspace{-0.35cm}
		\subfloat[\centering \emph{Mono (DS1)}]{\includegraphics[width=0.331\linewidth]{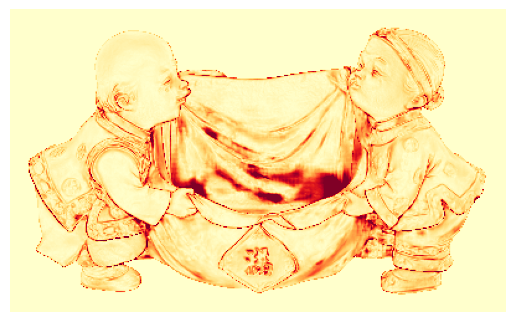}}\hspace{-0.06cm}
		\subfloat[\centering\emph{Multi\newline(DS1+DS2)}]{\includegraphics[width=0.331\linewidth]{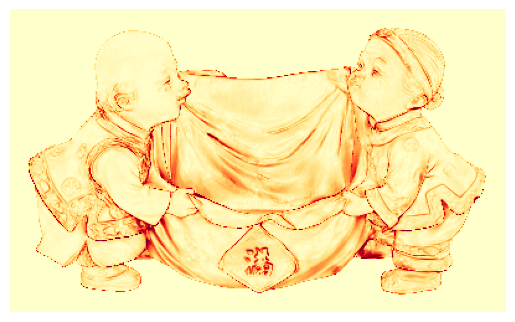}}\hspace{0.05cm}
		\subfloat[\centering \emph{Multi (DS1+DS2)}]{\includegraphics[width=0.291\linewidth]{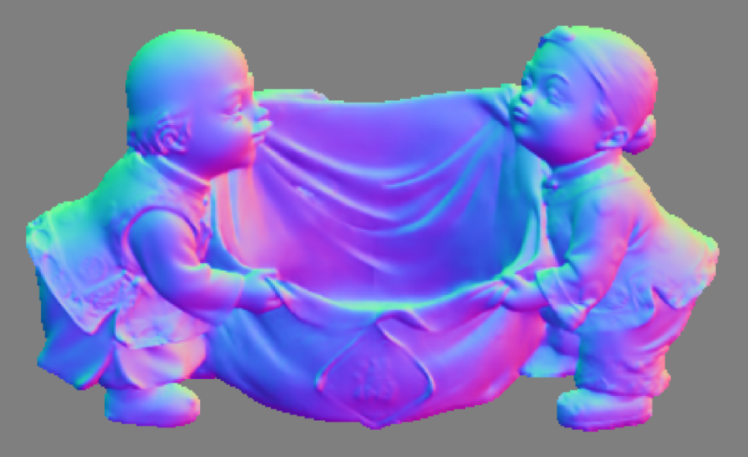}}
	\end{minipage}
	\vspace*{0.5em}
	\caption{Angular error map and predicted normal map for the ``reading'' and ``harvest'' objects from~\cite{dilidataset}. The concave parts, where cast shadows and inter-reflections occur, are better handled by our approach.}
	\label{fig:cast_shadow}
\end{figure}

Others common phenomenas which are cast-shadows and inter-reflections are also better handled by our multi-scale architecture, as Fig.~\ref{fig:cast_shadow} shows.

\begin{figure}[H]
	\centering
	\begin{subfigure}[b]{0.03\linewidth}
		\begin{turn}{90}
			\scriptsize
			~~~~~~~~\emph{Mono} (DS1)
		\end{turn}
	\end{subfigure}
	\begin{subfigure}[b]{0.29\linewidth}
		\includegraphics[width=\linewidth]{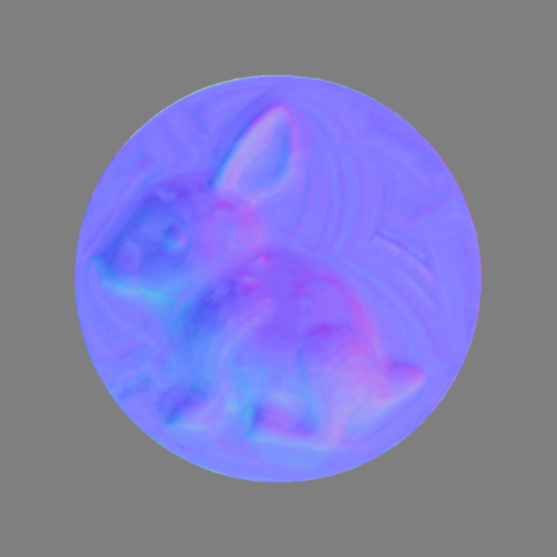}
	\end{subfigure}
	\begin{subfigure}[b]{0.29\linewidth}
		\includegraphics[width=\linewidth]{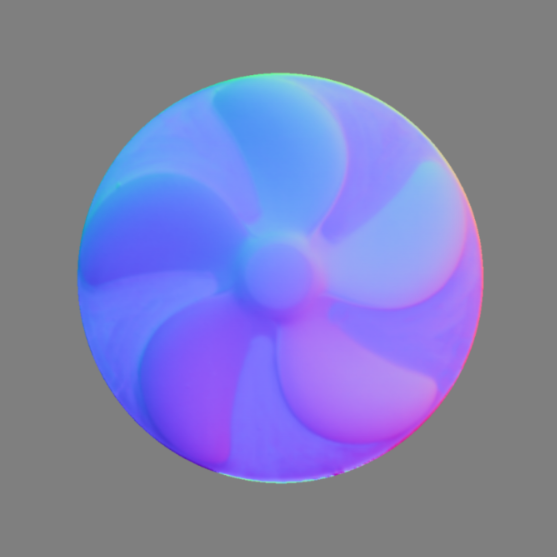}
	\end{subfigure}
	\begin{subfigure}[b]{0.29\linewidth}
		\includegraphics[width=\linewidth]{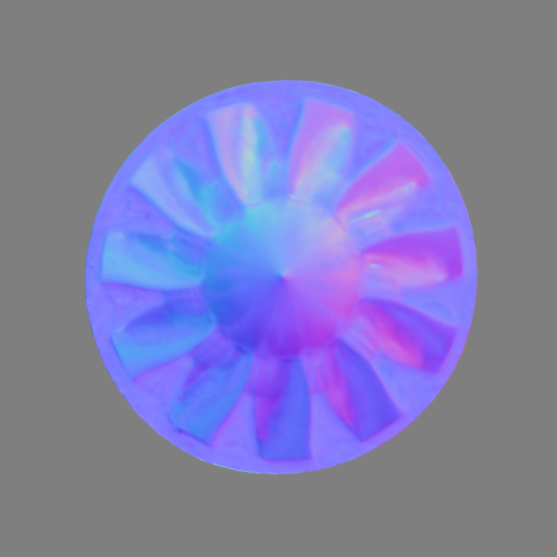}
	\end{subfigure} \hfill
	
	\vspace*{.25em}
	\begin{subfigure}[b]{0.03\linewidth}
		\begin{turn}{90}
			\scriptsize
			~~~~~~~~~~~~~\emph{Multi} (DS1+DS2)
		\end{turn}
	\end{subfigure}
	\begin{subfigure}[b]{0.29\linewidth}
		\includegraphics[width=\linewidth]{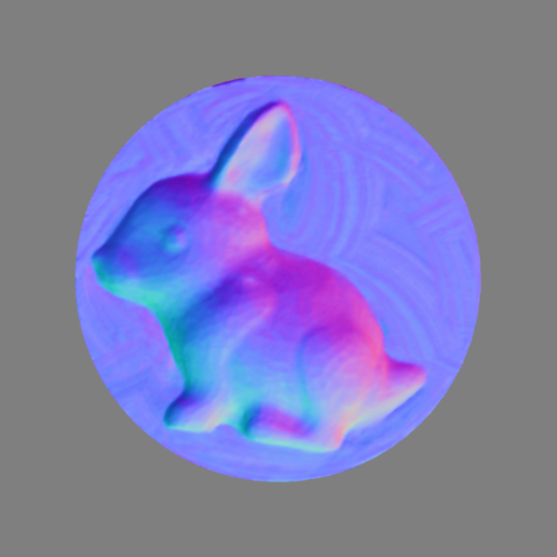}		\caption[]{\centering Bunny - Aluminium}
	\end{subfigure}
	\begin{subfigure}[b]{0.29\linewidth}
		\includegraphics[width=\linewidth]{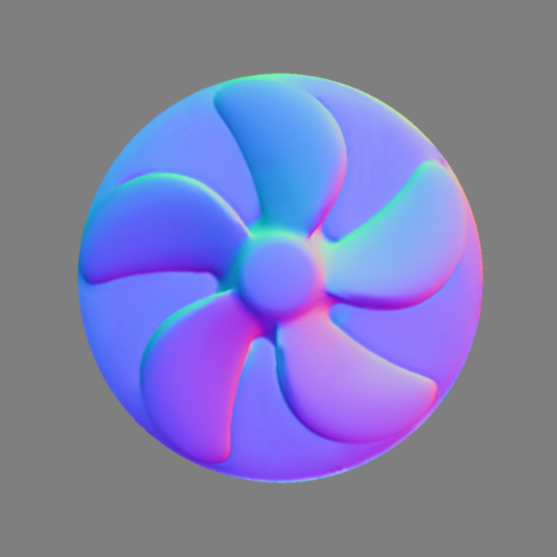}
		\caption[]{\centering Propeller - ABS}
	\end{subfigure}
	\begin{subfigure}[b]{0.29\linewidth}
		\includegraphics[width=\linewidth]{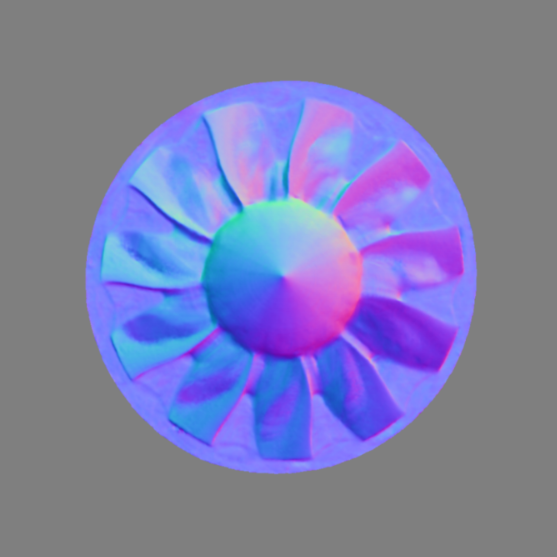}
		\caption[]{\centering Turbine - Steel}
		\label{fig:turbine}
	\end{subfigure} 
	\vspace*{0.6em}
	\caption{Visual comparison of the improvements brought by the combination of the new architecture and our new training set, on three objects from~\cite{DILI_10}. All three objects are much better reconstructed, although  the steel turbine remains challenging.}
	\label{fig:final_results_comparison}
\end{figure}

Fig.~\ref{fig:final_results_comparison} shows several additional qualitative comparisons of the result obtained with our baseline (mono-scale architecture trained on the existing dataset) and with both our building blocks included (multi-scale architecture trained on the new dataset). The convex objects (\emph{Bunny} and \emph{Propeller}) are very well reconstructed, despite being fabricated with anisotropic (Aluminium) or moderately specular (ABS, a type of plastic) materials. The steel turbine reconstruction is also improved, although on this object our approach shows its limitations. Indeed, this object exhibits concavities, which create many inter-reflections which are not very well handled by the network.

\begin{table*}[!ht]
	\scriptsize
	\resizebox{0.959\linewidth}{!}{
		\begin{tabular}{c|cccccccccc||c}
			& ball & bear & buddha & cat  & cow & goblet & harvest & pot1 & pot2 & reading & average \\
			\hline
			L2 (Baseline)\cite{1980} & 4.10 & 8.39 & 14.92 & 8.41 & 25.60 & 18.5 & 30.62 & 8.89 & 14.65 & 19.80 & 15.39 \\
			GPS-NET~\cite{GPS_net} & 2.92 & 5.07 & 7.77 & 5.42 & 6.14 & 9.00 & 15.14 & 6.04 & 7.01 & 13.58 & 7.81 \\
			CHR-PSN~\cite{Learning_conditional} & 2.26 & 6.35 & \underline{7.15} & 5.97 & 6.05 & 8.32 & 15.32 & 7.04 & 6.76 & 12.52 & 7.77 \\
			PS-transformer (10 images)~\cite{PS_transformer} & 3.27 & 4.88 & 8.65 & 5.34 & 6.54 & 9.28 &  14.41 & 6.06 & 6.97 & 11.24 & 7.66\\
			MT-PS-CNN~\cite{intra_inter} & 2.29 & 5.87 & \textbf{6.92} & 5.79 & 6.89 & \textbf{6.85} & \textbf{7.88} & 11.94 &7.48 & 13.71 &7.56 \\
			PS-FCN~\cite{9127824} & 2.67 & 7.72 & 7.52 & 4.75 & 6.72 & 7.84 & 12.39 &  6.17 & 7.15 & 10.92 & 7.39 \\
			CNN-PS~\cite{CNN_PS} & 2.2 & 4.6 & 7.9 & \underline{4.1} & 8.0 & 7.3 & 14.0 & 5.4 & 6.0 & 12.6 & 7.2 \\
			
			\textcolor{blue}{Mono (DS1)} & \textcolor{blue}{2.63} & \textcolor{blue}{6.66} & \textcolor{blue}{8.27} & \textcolor{blue}{4.47} & \textcolor{blue}{4.77} & \textcolor{blue}{8.24} & \textcolor{blue}{12.78} & \textcolor{blue}{6.00} & \textcolor{blue}{5.38} & \textcolor{blue}{9.68} & \textcolor{blue}{6.88} \\

			\textcolor{blue}{Multi (DS1)} & \textcolor{blue}{\textbf{1.60}} & \textcolor{blue}{7.82} & \textcolor{blue}{7.55} & \textcolor{blue}{4.33} & \textcolor{blue}{\underline{4.18}} & \textcolor{blue}{7.85} & \textcolor{blue}{12.36} & \textcolor{blue}{5.22} & \textcolor{blue}{5.36} & \textcolor{blue}{\underline{9.04}} & \textcolor{blue}{6.54} \\

			OB-Cnn~\cite{leveraging} & 2.49 & \underline{3.59} &  7.23 & 4.69 & 4.89 & \underline{6.89} & 12.79 & 5.10 & \textbf{4.98} & 11.08 & 6.37 \\
			PX-NET~\cite{PX_NET} &  \underline{2.03} & \textbf{3.58} & 7.61 & 4.39 & 4.69 & 6.90 & 13.10 & \underline{5.08} & 5.10 & 10.26 & \underline{6.28}\\
			
			\textcolor{blue}{Multi (DS1+DS2)} & \textcolor{blue}{2.05} & \textcolor{blue}{4.24} & \textcolor{blue}{\underline{7.03}} & \textcolor{blue}{\textbf{3.9}} & \textcolor{blue}{\textbf{4.00}} & \textcolor{blue}{7.57} & \textcolor{blue}{\underline{11.01}} & \textcolor{blue}{\textbf{4.94}} & \textcolor{blue}{\underline{5.22}} & \textcolor{blue}{\textbf{8.47}} & \textcolor{blue}{\textbf{5.84}} \\
		\end{tabular}
	}
\vspace*{0.5em}
	\caption{Mean angular error (in degrees) on the DiLiGenT~\cite{dilidataset} benchmark.The best result for each object is indicated in bold, and the second best one is underlined. The lines in blue indicate our results. Combining the proposed multi-scale architecture ``\emph{Multi}'' and proposed training dataset ``\emph{DS2}'' yields state-of-the-art results, by a large margin.}
	\label{tab:comparison_mono_multi_digi}
\end{table*}

\subsection{Quantitative evaluation on DiLiGenT~\cite{dilidataset}}

Next, we compare in Table~\ref{tab:comparison_mono_multi_digi} our results against the most recent state-of-the-art methods, on the DiLiGenT benchmark~\cite{dilidataset}. Let us however remark that PS-transformer~\cite{PS_transformer} takes as inputs no more than 10 images, hence the comparison is biased. Besides, we emphasize that our mono-scale architecture is largely inspired from PS-FCN~\cite{PS_FCN,9127824}, hence \emph{Mono} (DS1) can be considered as an optimized version of~\cite{PS_FCN,9127824}, where we let the training phase run for much longer. This table shows that the proposed multi-scale architecture provides a significant gain of $4.6\%$, in comparison with the mono-scale approach -- compare \emph{Mono} (DS1) and \emph{Multi} (DS1). And, as soon as our new training dataset is considered, the state-of-the-art is outperformed and we reach an average angular error below $6^\circ$, with a particularly visible improvement on the most difficult ``reading'' object (Fig.~\ref{fig:cast_shadow}).

\subsection{Quantitative evaluation on DiLiGenT $10^2$~\cite{DILI_10}}

We now quantatively evaluate the impact of the multi-scale architecture on the DiLiGenT $10^2$ benchmark~\cite{DILI_10}. Note that we process images at their full resolution (1024 pixels by 1024), requiring 7 scales in the multi-scale architecture. To this end, we show in Table~\ref{tab:comparison_mono_multi_digi_10} the difference between the mono- and the multi-scale approaches, when they are both trained on the pre-existing dataset. As can be observed, a significant gain of $9.3\%$ is observed with the multi-scale architecture. The gain is most visible on objects which have a spherical shape and anisotropic material (top right of Tab~\ref{tab:difference_mono_multi}, see also Fig.~\ref{fig:mono_vs_multi} for a qualitative result on the Golf - CU object), as well as for the most challenging ``acrylic'' material, which is translucent.

\begin{table}[H]
	\centering
	\begin{subtable}{.49\linewidth}\centering
		\includegraphics[width=\linewidth]{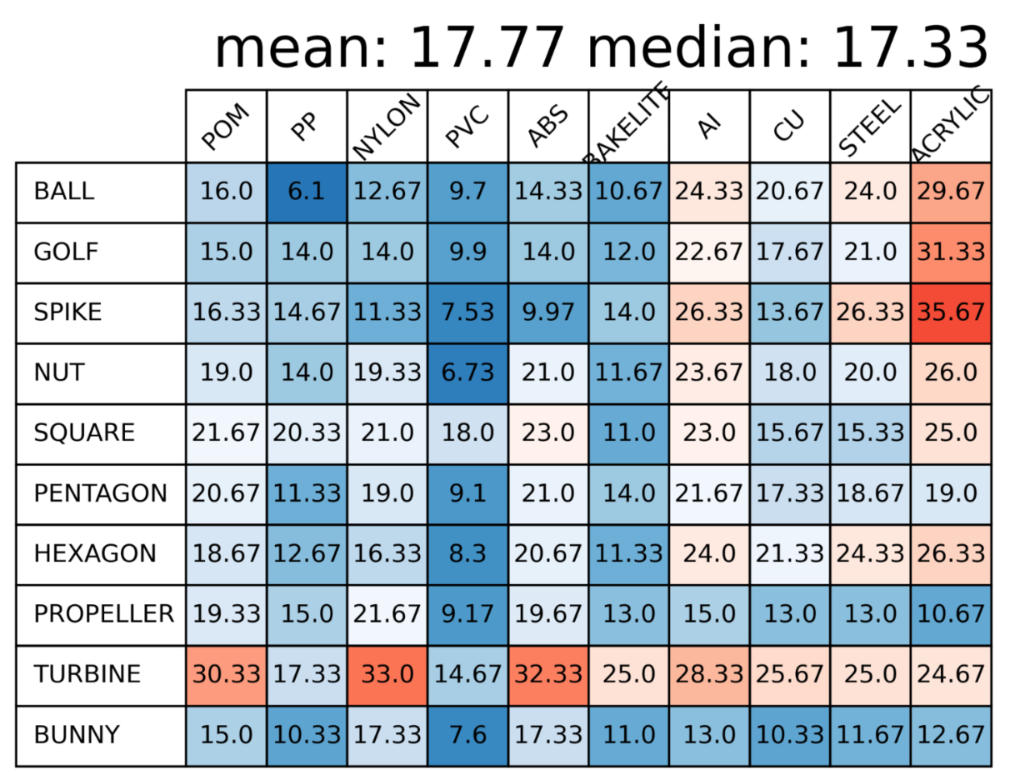}
		\caption{\centering \emph{Mono} (DS1)}\label{tab:mean_mono}
	\end{subtable}
	\begin{subtable}{.49\linewidth}\centering
		\includegraphics[width=\linewidth]{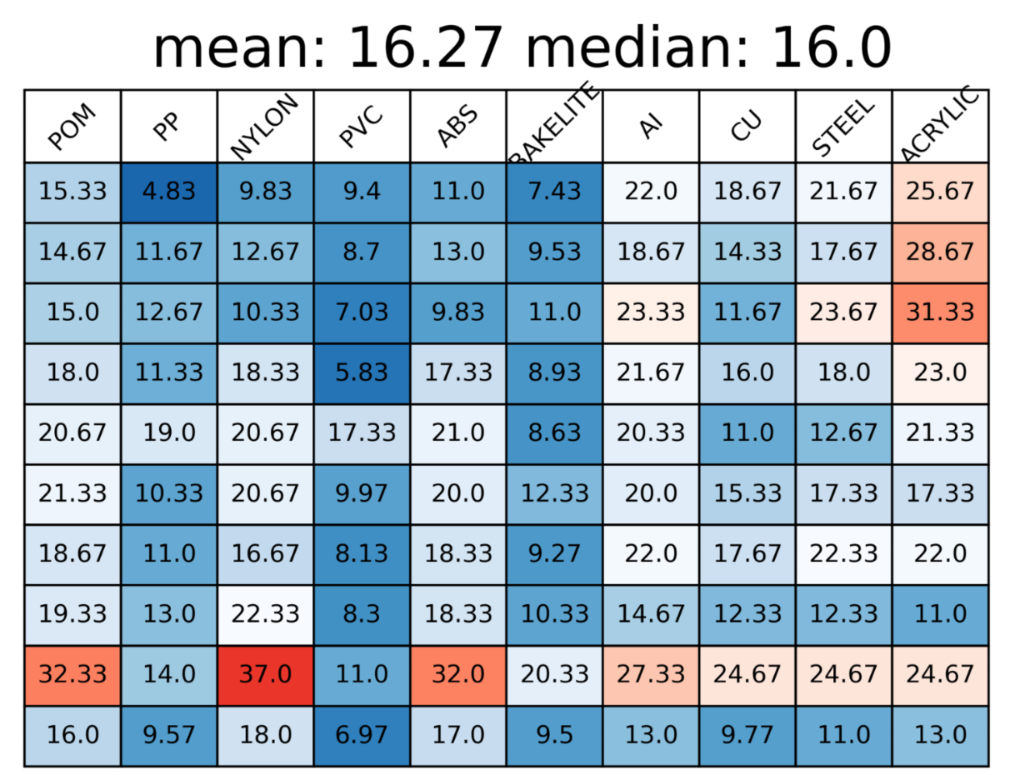}
		\caption{\centering \emph{Multi} (DS1)}\label{tab:mean_multi}
	\end{subtable} ~ \\
	\begin{subtable}{.48\linewidth}\centering
		\includegraphics[width=\linewidth]{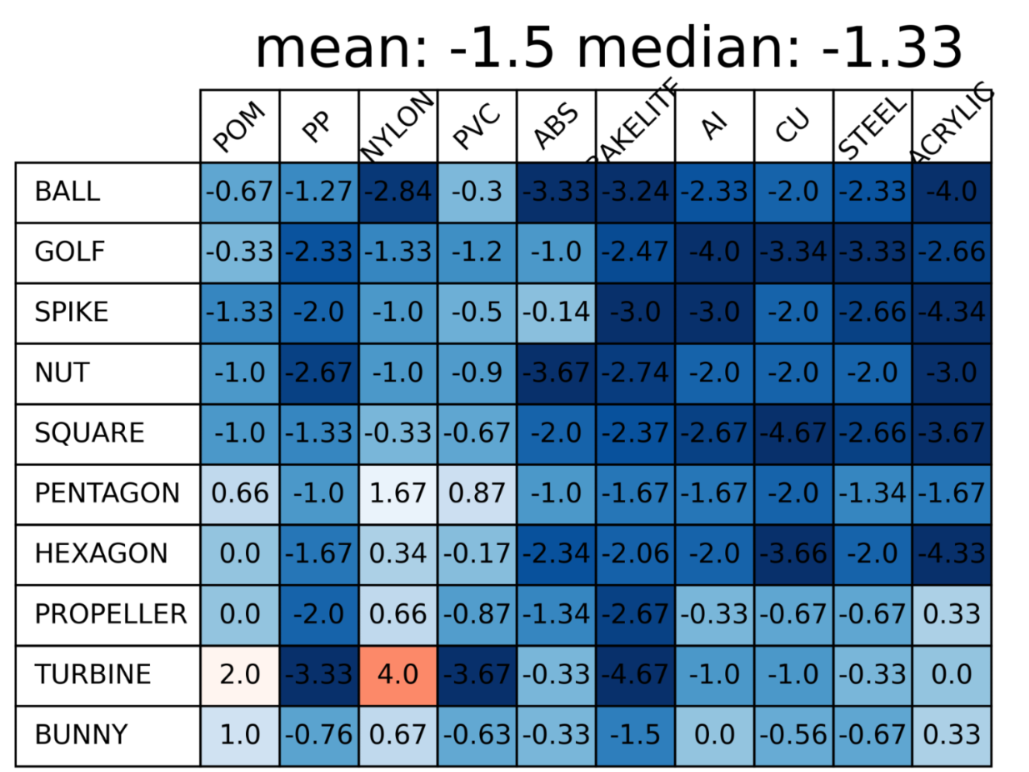}
		\caption{\scriptsize \emph{Multi} (DS1) - \emph{Mono} (DS1)}\label{tab:difference_mono_multi}
	\end{subtable}\vspace*{0.75em}
	\caption{Mean angular on the $\emph{DiLiGenT10}^2$ benchmark, considering either the mono-scale architecture or the multi-scale one, both trained on the pre-existing dataset DS1. The multi-scale approach yields a significant gain, most visible on the top-right part of the table (spherical shapes with anisotropic reflectance).}
	\label{tab:comparison_mono_multi_digi_10}
\end{table}

\begin{table*}
	\centering
	\begin{subtable}{.32\linewidth}\centering
		\includegraphics[height=0.75\linewidth]{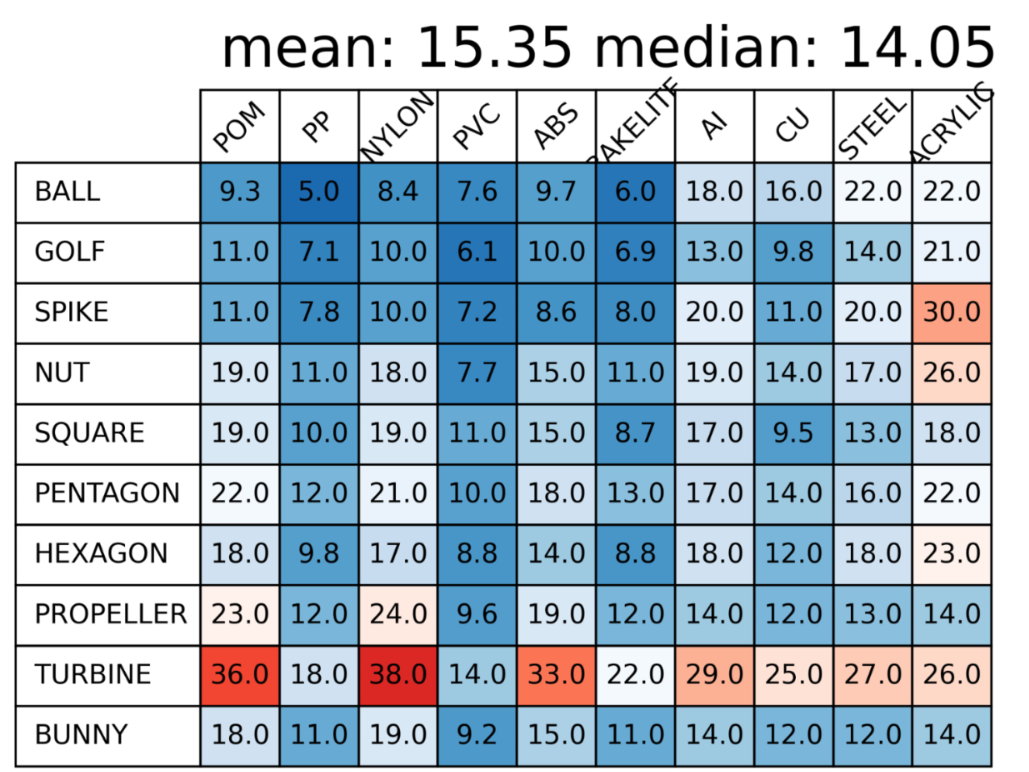}
		\caption{\centering \emph{Mono} (DS1+DS2)}\label{tab:multi_new_data_patches}
	\end{subtable}
	\begin{subtable}{.32\linewidth}\centering
		\includegraphics[height=0.75\linewidth]{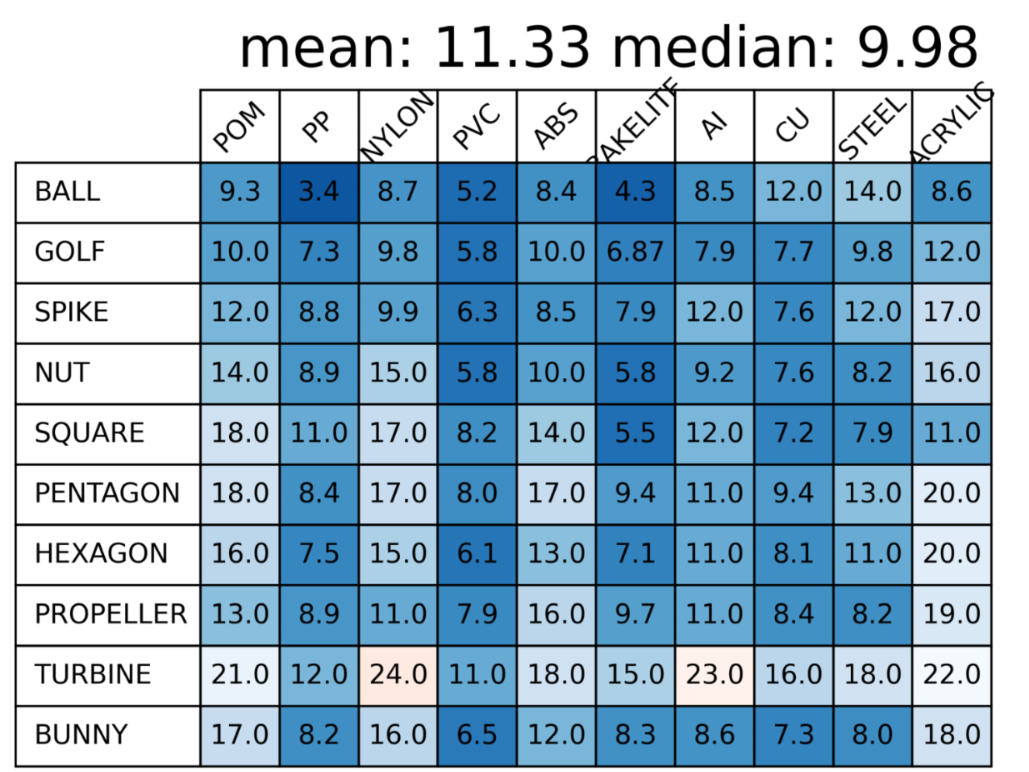}
		\caption{\centering \emph{Multi} (DS1+DS2)}\label{tab:multi_new_data_full}
	\end{subtable}
	\begin{subtable}{.32\linewidth}\centering
		\includegraphics[height=0.75\linewidth]{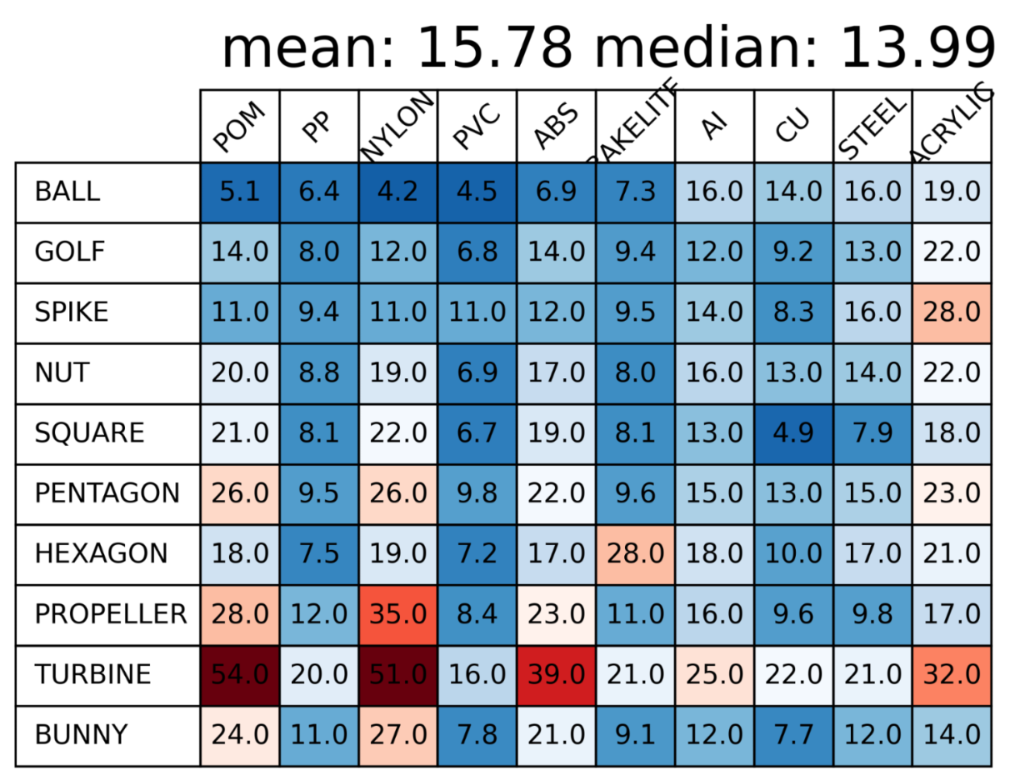}
		\caption{\centering CNN-PS~\cite{CNN_PS} (DS1)}\label{tab:dili10_cnn_ps}
	\end{subtable} \vspace*{1em}
	\caption{Mean angular error on the $\emph{DiLiGenT10}^2$ benchmark, with the results of CNN-PS~\cite{CNN_PS} indicated for comparison. When incorporating both the new dataset and the multi-scale architecture, the state-of-the-art is largely outperformed.}\vspace*{-.5em}
	\label{tab:multi_new_data}
\end{table*}

We repeat this experiment in Table~\ref{tab:multi_new_data}, but this time with our networks trained on the new dataset. Comparing Tables~\ref{tab:comparison_mono_multi_digi_10} and~\ref{tab:multi_new_data} allows one to quantify the benefits of using our new training dataset: the mono-scale architecture gets improved by $14\%$, and the multi-scale one by $30\%$. Comparing Tables~\ref{tab:multi_new_data_patches} and~\ref{tab:multi_new_data_full} also allows one to quantify the impact of switching to the multi-scale architecture: the results improve by $26\%$. Particularly large improvements can be observed on the \emph{Turbine} and \emph{Acrylic Gulf} objects (see also Figs.~\ref{fig:turbine} and~\ref{fig:zoom_acrylique_translucide}). For such objects with non-local light transport (due to inter-reflections or anisotropic reflectance), the ability of tne multi-scale approach to get access to a global information is indeed of primary importance. 

Overall, the combination of the new architecture and dataset allows one to reach an average error of $11.33^\circ$ on this benchmark. This is to be compared with the $15.78^\circ$ achieved by CNN-PS~\cite{CNN_PS} (Table~\ref{tab:dili10_cnn_ps}), which was the best performing method so far~\cite{DILI_10}. By comparing our results with all available state-of-the-art methods ~\cite{SDPS_net,PS_FCN,CNN_PS,Dili_PF14,Dili_ST14,Dili_ls,IRPS,Dili_WG10,GPS_net,spline_net}, we found out that the proposed method is the best performer on $73\%$ of the objects of this benchmark, as indicated in Table~\ref{tab:our_vs_world}.

\begin{table}[H]
	\centering
	\includegraphics[width=0.65\linewidth]{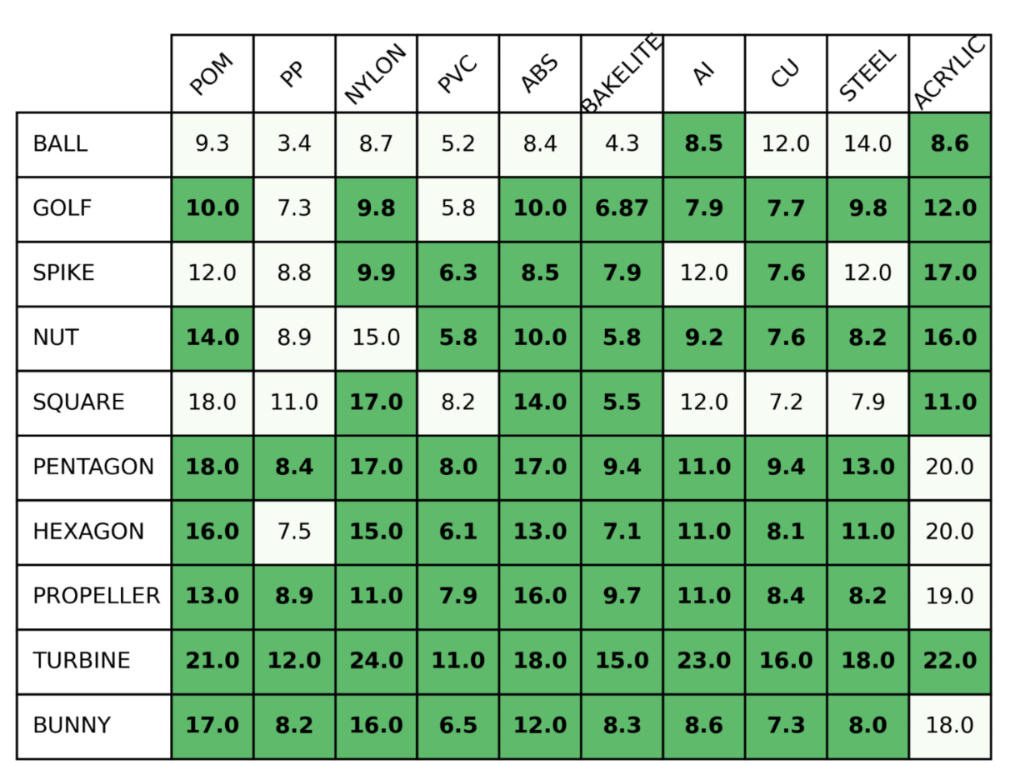}\vspace*{.75em}
	\caption{Mean angular error achieved by the best performer among~\cite{SDPS_net,PS_FCN,CNN_PS,Dili_PF14,Dili_ST14,Dili_ls,IRPS,Dili_WG10,GPS_net,spline_net} and us, on the $100$ objects of~\cite{DILI_10}. Green cases indicate when the proposed architecture, combined with the new dataset, gives the best results. \vspace*{-0.2em}}\label{tab:our_vs_world}
\end{table}

\subsection{Limitations}
\begin{figure}[!ht]
	\centering
	\captionsetup[subfigure]{labelformat=empty}
	\subfloat[\centering{CNN-PS} (DS1) Error: 32$\degree$]{\includegraphics[width=0.301\linewidth]{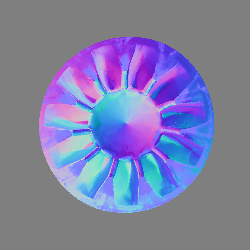}}\hspace{0.05cm}
	\subfloat[\emph{Mono} (DS1)\\ \centering Error: 26$\degree$]{\includegraphics[width=0.301\linewidth]{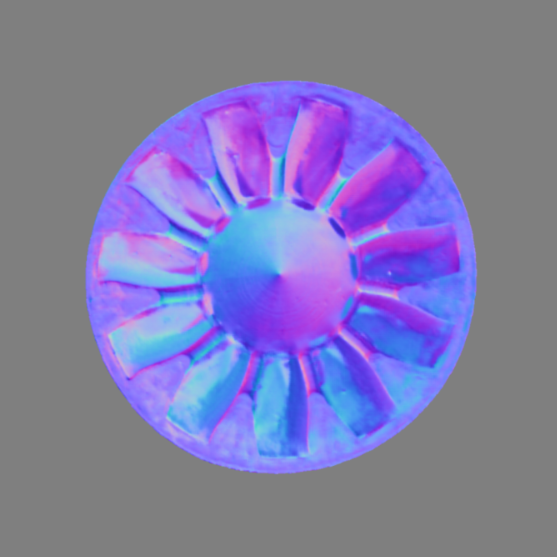}}\hspace{0.05cm}
	\subfloat[\centering \emph{Multi} (DS1+DS2) Error: 22$\degree$]{\includegraphics[width=0.301\linewidth]{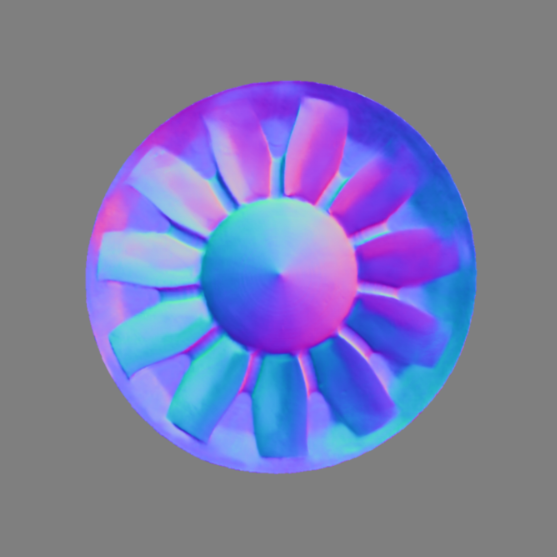}}\hspace{0.05cm}
	\vspace*{0.8em}
	\caption{Results of CNN-PS, our mono-scale and our multi-scale architecture on the acrylic turbine.}
	\label{fig:limitation_pred}
\end{figure}

\begin{figure}[!ht]
	\centering
	\captionsetup[subfigure]{labelformat=empty}
	\subfloat[ABS]{\includegraphics[width=0.401\linewidth]{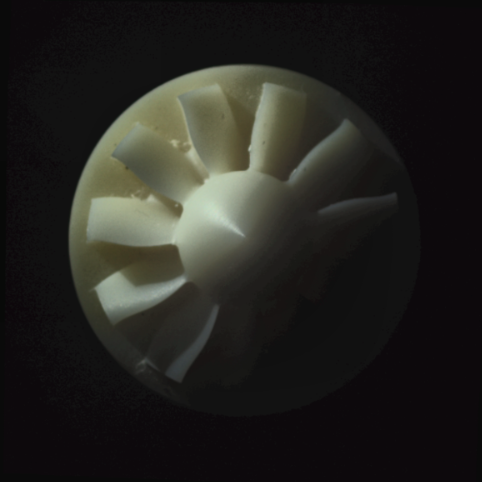}}\hspace{0.05cm}
	\subfloat[Acrylic]{\includegraphics[width=0.401\linewidth]{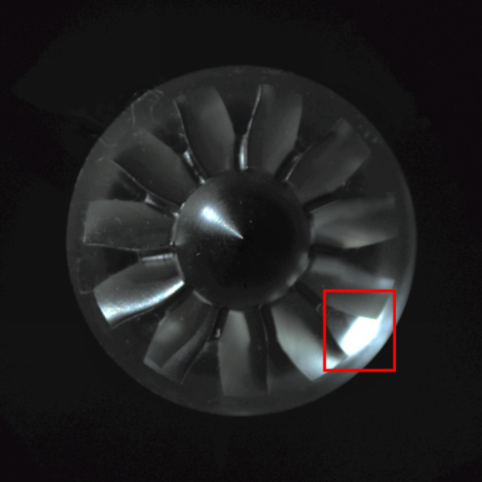}}\hspace{0.05cm}
	\vspace*{0.8em}
	\caption{Same turbine, fabricated either with a diffuse (ABS) or an anisotropic (acrylic) material, and illuminated from the same direction (coming from ``top left''). The bottom-right area, which is shadowed in the diffuse case, appears much shinier on the anisotropic object.}
	\label{fig:limitation_inputs}
	\end{figure}

Even if the combination of our multi-scale and our new training dataset improves the results on non-Lambertian materials, some shortcomings remain. For example, we notice that the normals at the border of some translucent objects are incorrectly predicted (Fig.~\ref{fig:limitation_pred}). As shown in Fig.~\ref{fig:limitation_inputs}, in this example the the opposite side of the incoming light is the most shiny part of the image. Although our multi-scale approach better handles such anisotropic than the mono-scale one or existing methods such as CNN-PS, it shows its limitations when the anisotropy is this much important. 

\section{Conclusion}

In this paper, we have proposed a novel deep normal estimation framework for the calibrated photometric stereo problem. It builds upon a multi-scale architecture which is independent from the resolution of the images, as well as a new comprehensive learning dataset. We have shown on publicly available benchmarks that the combination of these two features yields state-of-the-art results, with performances particularly improved on challenging anisotropic materials. In the future, we plan to extend our approach to handle observation maps~\cite{CNN_PS} as well, which have recently been shown to benefit from physical interpretability~\cite{ICIP_2022}.

\paragraph{Acknowledgment}
This work was granted access to the HPC resources of IDRIS under the allocation 2022-AD010613775 made by GENCI.

\footnotesize
\bibliographystyle{plain}
\bibliography{biblio.bib}

\begin{thebibliography}{10}

\bibitem{Ambientcg}
{AmbientCG}.
\newblock \url{https://ambientcg.com/}.

\bibitem{Sketchfab}
{Sketchfab}.
\newblock \url{https://sketchfab.com}.

\bibitem{burley2012physically}
B.~Burley and W.~D. Studios.
\newblock {Physically-based shading at Disney}.
\newblock {\em ACM SIGGRAPH Courses}, 2012.

\bibitem{intra_inter}
Y.~Cao, B.~Ding, Z.~He, J.~Yang, J.~Chen, Y.~Cao, and X.~Li.
\newblock {Learning inter-and intraframe representations for non-Lambertian
  photometric stereo}.
\newblock {\em OLEN}, 150:106838, 2022.

\bibitem{SDPS_net}
G.~Chen, K.~Han, B.~Shi, Y.~Matsushita, and Kwan-Yee~K. Wong.
\newblock {Self-Calibrating Deep Photometric Stereo Networks}.
\newblock In {\em CVPR}, 2019.

\bibitem{PS_FCN}
G.~Chen, K.~Han, and K.~Wong.
\newblock {PS-FCN: A Flexible Learning Framework for Photometric Stereo}.
\newblock In {\em ECCV}, 2018.

\bibitem{9127824}
G.~Chen, Kai Han, Boxin S., Yasuyuki M., and K.~W.
\newblock {Deep Photometric Stereo for Non-Lambertian Surfaces}.
\newblock {\em PAMI}, 44(1), 2022.

\bibitem{blender}
Blender~Online Community.
\newblock {\em {Blender - a 3D modelling and rendering package}}, 2018.

\bibitem{haefner2019variational}
B.~Haefner, Z.~Ye, M.~Gao, T.~Wu, Y.~Qu{\'e}au, and D.~Cremers.
\newblock Variational uncalibrated photometric stereo under general lighting.
\newblock In {\em ICCV}, 2019.

\bibitem{leveraging}
D.~Honz{\'a}tko, E.~T{\"u}retken, P.~Fua, and L.~Dunbar.
\newblock {Leveraging Spatial and Photometric Context for Calibrated
  Non-Lambertian Photometric Stereo}.
\newblock In {\em 3DV}, 2021.

\bibitem{CNN_PS}
S.~Ikehata.
\newblock {CNN-PS: CNN-based Photometric Stereo for General Non-Convex
  Surfaces}.
\newblock In {\em ECCV}, 2018.

\bibitem{PS_transformer}
S.~Ikehata.
\newblock {PS-transformer: Learning sparse photometric stereo network using
  self-attention mechanism}.
\newblock In {\em BMVC}, 2021.

\bibitem{ICIP_2022}
S.~Ikehata.
\newblock {Does Physical Interpretability of Observation Map Improve
  Photometric Stereo Networks?}
\newblock In {\em ICIP}, 2022.

\bibitem{uni_ps}
S.~Ikehata.
\newblock Universal photometric stereo network using global lighting contexts.
\newblock {\em CVPR}, 2022.

\bibitem{blobby_dataset}
M.~Johnson and E.~Adelson.
\newblock {Shape Estimation in Natural Illumination}.
\newblock In {\em CVPR}, 2011.

\bibitem{9376632}
Y.~Ju, J.~Dong, and S.~Chen.
\newblock {Recovering Surface Normal and Arbitrary Images: A Dual Regression
  Network for Photometric Stereo}.
\newblock {\em TIP}, 30:3676--3690, 2021.

\bibitem{9301860}
Y.~Ju, M.~Jian, J.~Dong, and K.~Lam.
\newblock {Learning Photometric Stereo via Manifold-based Mapping}.
\newblock In {\em VCIP}, 2020.

\bibitem{pay_attention}
Y.~Ju, K.~Lam, Y.~Chen, L.~Qi, and J.~Dong.
\newblock {Pay Attention to Devils: A Photometric Stereo Network for Better
  Details}.
\newblock In {\em IJCAI}, 2020.

\bibitem{Learning_conditional}
Y.~Ju, Y.~Peng, M.~Jian, F.~Gao, and J.~Dong.
\newblock Learning conditional photometric stereo with high-resolution
  features.
\newblock {\em CVM}, 8(1):105--118, 2022.

\bibitem{uncalibrated_neural_inverse}
B.~Kaya, S.~Kumar, C.~Oliveira, V.~Ferrari, and Van G.
\newblock {Uncalibrated Neural Inverse Rendering for Photometric Stereo of
  General Surfaces}.
\newblock In {\em CVPR}, 2021.

\bibitem{adam}
Diederik~P Kingma and Jimmy Ba.
\newblock Adam: a method for stochastic optimization.
\newblock In {\em ICLR}, 2015.

\bibitem{minify}
J.~Li, A.~Robles-Kelly, S.~You, and Y.~Matsushita.
\newblock {Learning to Minify Photometric Stereo}.
\newblock In {\em CVPR}, 2019.

\bibitem{fast_near}
D.~Lichy, S.~Sengupta, and D.~Jacobs.
\newblock Fast light-weight near-field photometric stereo.
\newblock In {\em CVPR}, 2022.

\bibitem{shape_and_material}
D.~Lichy, J.~Wu, S.~Sengupta, and D.~Jacobs.
\newblock {Shape and Material Capture at Home}.
\newblock In {\em CVPR}, 2021.

\bibitem{PX_NET}
F.~Logothetis, I.~Budvytis, R.~Mecca, and R.~Cipolla.
\newblock {PX-net: Simple and efficient pixel-wise training of photometric
  stereo networks}.
\newblock In {\em ICCV}, 2021.

\bibitem{logothetis2022cnn}
F.~Logothetis, R.~Mecca, I.~Budvytis, and R.~Cipolla.
\newblock A {CNN} based approach for the point-light photometric stereo
  problem.
\newblock {\em IJCV}, 131(1):101--120, 2023.

\bibitem{marching}
W.~Lorensen and H.~Cline.
\newblock {Marching cubes: A high resolution 3D surface construction
  algorithm}.
\newblock {\em SIGGRAPH}, 1987.

\bibitem{MERL}
W.~Matusik.
\newblock {\em A data-driven reflectance model}.
\newblock PhD thesis, Massachusetts Institute of Technology, 2003.

\bibitem{mo2018uncalibrated}
Z.~Mo, B.~Shi, F.~Lu, S.-K. Yeung, and Y.~Matsushita.
\newblock Uncalibrated photometric stereo under natural illumination.
\newblock In {\em CVPR}, 2018.

\bibitem{Dili_PF14}
T.~Papadhimitri and P.~Favaro.
\newblock {A Closed-Form, Consistent and Robust Solution to Uncalibrated
  Photometric Stereo Via Local Diffuse Reflectance Maxima}.
\newblock {\em IJCV}, 2014.

\bibitem{DILI_10}
J.~Ren, F.~Wang, J.~Zhang, Q.~Zheng, M.~Ren, and B.~Shi.
\newblock {DiLiGenT10{$^2$}: A Photometric Stereo Benchmark Dataset with
  Controlled Shape and Material Variation}.
\newblock In {\em CVPR}, 2022.

\bibitem{santo2020deep}
H.~Santo, M.~Waechter, and Y.~Matsushita.
\newblock Deep near-light photometric stereo for spatially varying
  reflectances.
\newblock In {\em ECCV}, 2020.

\bibitem{Dili_ST14}
B.~Shi, P.~Tan, Y.~Matsushita, and K.~Ikeuchi.
\newblock Bi-polynomial modeling of low-frequency reflectances.
\newblock {\em PAMI}, 36(6):1078--1091, 2013.

\bibitem{dilidataset}
B.~Shi, Z.~Wu, Z.~Mo, D.~Duan, S.~Yeung, and P.~Tan.
\newblock {A Benchmark Dataset and Evaluation for Non-Lambertian and
  Uncalibrated Photometric Stereo}.
\newblock In {\em CVPR}, 2016.

\bibitem{Dili_ls}
B.~Shi, Z.~Wu, Z.~Mo, D.~Duan, S.~Yeung, and P.~Tan.
\newblock {A benchmark dataset and evaluation for non-lambertian and
  uncalibrated photometric stereo}.
\newblock In {\em CVPR}, 2016.

\bibitem{IRPS}
T.~Taniai and T.~Maehara.
\newblock {Neural Inverse Rendering for General Reflectance Photometric
  Stereo}.
\newblock In {\em ICML}, 2018.

\bibitem{9069410}
X.~Wang, Z.~Jian, and M.~Ren.
\newblock {Non-Lambertian Photometric Stereo Network Based on Inverse
  Reflectance Model With Collocated Light}.
\newblock {\em TIP}, 29, 2020.

\bibitem{1980}
R.~J. Woodham.
\newblock {Photometric Method For Determining Surface Orientation From Multiple
  Images}.
\newblock {\em Opt. Eng.}, 19, 1980.

\bibitem{Dili_WG10}
L.~Wu, A.~Ganesh, B.~Shi, Y.~Matsushita, Y.~Wang, and Y.~Ma.
\newblock Robust photometric stereo via low-rank matrix completion and
  recovery.
\newblock In {\em ACCV}, 2011.

\bibitem{GPS_net}
Z.~Yao, K.~Li, Y.~Fu, H.~Hu, and B.~Shi.
\newblock {GPS-Net: Graph-based Photometric Stereo Network}.
\newblock In {\em NIPS}, 2020.

\bibitem{spline_net}
Q.~Zheng, Y.~Jia, B.~Shi, X.~Jiang, L.~Duan, and A.~Kot.
\newblock {SPLINE-Net: Sparse Photometric Stereo Through Lighting Interpolation
  and Normal Estimation Networks}.
\newblock In {\em ICCV}, 2019.

\end{thebibliography}

\end{document}